\documentclass[10pt,journal,compsoc]{IEEEtran}
\ifCLASSOPTIONcompsoc
\usepackage[nocompress]{cite}
\else
\usepackage{cite}
\fi

\ifCLASSINFOpdf
\else
\fi

\usepackage{amsmath}
\usepackage{algorithmic}
\usepackage{graphicx}
\usepackage[dvipsnames]{xcolor}
\usepackage{amssymb}
\usepackage{mathrsfs}
\usepackage{multirow}
\usepackage{array}
\usepackage{url}
\usepackage{wrapfig}
\usepackage{xspace}
\usepackage{booktabs}

\usepackage{bm}
\usepackage{siunitx}
\usepackage[super]{nth}

\usepackage[pagebackref=true,breaklinks=true,letterpaper=true,colorlinks,linkcolor=blue,citecolor=blue,bookmarks=false]{hyperref}

\usepackage{bbm}
\usepackage{colortbl}
\usepackage{lipsum}

\usepackage{colortbl} %
\definecolor{mygray-bg}{gray}{0.9}

\usepackage[short]{optidef}

\usepackage{svg}

\makeatletter
\DeclareRobustCommand\onedot{\futurelet\@let@token\@onedot}
\def\@onedot{\ifx\@let@token.\else.\null\fi\xspace}

\def\eg{\emph{e.g}\onedot} 
\def\ie{\emph{i.e}\onedot} 
 
 \def\vs{\emph{vs}\onedot}

\makeatother

\makeatletter

\newcommand{\Rmnum}[1]{\expandafter\@slowromancap\romannumeral #1@}
\makeatother

\newcommand{\revise}[1]{\textcolor{black}{#1}}
\newcommand{\Revise}{\color{black}}

\newcommand{\minisection}[1]{\noindent\textbf{#1}}

\begin{document}
	\title{SMPLer: Taming Transformers for Monocular 3D Human Shape and Pose Estimation}
	\author{Xiangyu~Xu,~ %
		Lijuan~Liu,~ Shuicheng Yan
		\IEEEcompsocitemizethanks{
			\IEEEcompsocthanksitem
			X. Xu is with Xi'an Jiaotong University, Xi'an, China. E-mail: xuxiangyu2014@gmail.com.
			\IEEEcompsocthanksitem L. Liu is with Sea AI Lab, Singapore. E-mail: liulj@sea.com.
			\IEEEcompsocthanksitem S. Yan is with Skywork AI, Singapore. E-mail: shuicheng.yan@gmail.com.
		}%
	}

	\markboth{IEEE Transactions on Pattern Analysis and Machine Intelligence}{}

	\IEEEtitleabstractindextext{%
		\begin{abstract}
			Existing Transformers for monocular 3D human shape and pose estimation typically have a quadratic computation and memory complexity with respect to the feature length, which hinders the exploitation of fine-grained information in high-resolution features that is beneficial for accurate reconstruction.
			In this work, we propose an \textbf{SMPL}-based Transform\textbf{er} framework (SMPLer) to address this issue.
			SMPLer incorporates two key ingredients: a decoupled attention operation and an SMPL-based target representation, which allow effective utilization of high-resolution features in the Transformer.
			In addition, based on these two designs, we  also introduce several novel modules including a multi-scale attention and a joint-aware attention to further boost  the reconstruction performance.
			Extensive experiments demonstrate the effectiveness of SMPLer against existing 3D human shape and pose estimation methods both quantitatively and qualitatively.
			Notably, the proposed algorithm achieves an MPJPE of 45.2 mm on the Human3.6M dataset, improving upon Mesh Graphormer by more than 10\% with fewer than one-third of the parameters.
			Code and pretrained models are available at \url{https://github.com/xuxy09/SMPLer}.
		\end{abstract}

		}

	\maketitle

	\IEEEdisplaynontitleabstractindextext
	\IEEEpeerreviewmaketitle

	\begin{figure*}[t]
		\footnotesize
		\begin{center}
			\begin{tabular}{c}
				\hspace{0mm}
				\includegraphics[width = 0.95\linewidth]{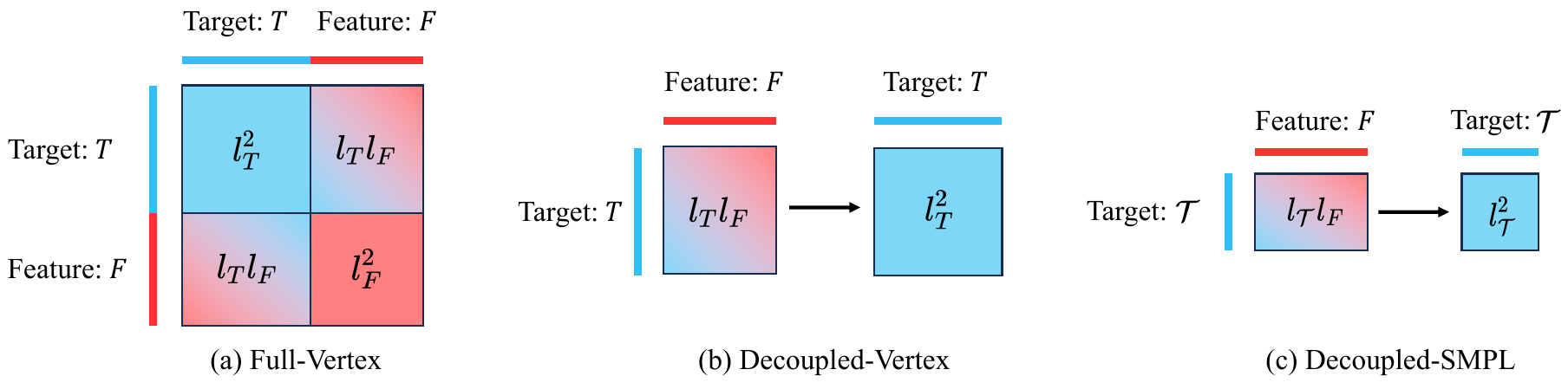} 
			\end{tabular}
		\end{center}
		\vspace{-3mm}
		\caption{Two key designs of the proposed Transformer. The sub-caption ``A-B''  denotes the attention form ``A'' and the target representation ``B'', respectively. 
		The vertical and horizontal lines around the rectangles represent query and key in the attention operation.
		Red indicates source image features, blue indicates target output representation, and
		the colors within the rectangles represent the interactions between them.
			(a) Existing Transformers for 3D human reconstruction~\cite{lin2021end,lin2021mesh} typically adopt a ViT-style full attention operation and a vertex-based target representation, hindering the utilization of high-resolution image features.
			In contrast, we propose a decoupled attention (b) and an SMPL-based target representation (c), which effectively address the above problem and improve reconstruction performance.
			$l_T$, $l_\mathcal{T}$, and $l_F$ are the lengths of the vertex-based embedding, SMPL-based embedding, and image features, respectively.
			The area of each rectangle denotes the computation and memory complexity of the attention operation.
			\revise{Please refer to Section~\ref{sec: efficient attention} for mathematical explanations.}
		}
		\vspace{-2mm}
		\label{fig: teaser}
	\end{figure*}
	
	\IEEEraisesectionheading{
		\section{Introduction}\label{sec:intro}}
	\IEEEPARstart{M}{onocular} 3D human shape and pose estimation is a fundamental task in computer vision, which aims to recover the unclothed human body shape as well as its 3D pose from a single input image~\cite{bogo2016keep,kanazawa2018end,kocabas2019vibe,kolotouros2019spin,kolotouros2019convolutional}. 
	It has been widely used in many applications including visual tracking~\cite{rajasegaran2021tracking,rajasegaran2022tracking}, virtual/augmented reality~\cite{wang2019re,xu2021texformer,zheng2020pamir,mir2020learning,ma2021power,alldieck2019learning,alldieck2018video,alldieck2019tex2shape,alldieck2018detailed}, motion generation~\cite{li2021ai,hong2022avatarclip}, image manipulation~\cite{sanyal2021learning,grigorev2021stylepeople}, and neural radiance field~\cite{peng2021neural,kwon2021neural,chen2022gpnerf}.
	Different from multi-view 3D reconstruction where the solution is well restricted by geometrical constraints~\cite{hartley2003multiple},  
	this task is particularly challenging due to the inherent depth ambiguity of a single 2D image, and usually requires strong prior knowledge learned from large amounts of data to generate plausible results.
	Thus, the state-of-the-art algorithms~\cite{lin2021end,lin2021mesh} use Transformers~\cite{vaswani2017attention} for this task due to their powerful capabilities of grasping knowledge and learning representations from data~\cite{devlin2018bert,dosovitskiy2020image}.

	Existing Transformers for monocular 3D human shape and pose estimation~\cite{lin2021end,lin2021mesh} generally follow the ViT style~\cite{dosovitskiy2020image} to design the network.
	As shown in Figure~\ref{fig: teaser}(a), the target embeddings are first concatenated with the input features and then processed by a full attention layer that models all pairwise dependencies including target-target, target-feature, feature-target, and feature-feature.
	While this design has achieved impressive results, it leads to quadratic computation and memory complexity with respect to the length of the image feature, \ie $\mathcal{O}((l_F+l_T)^2)$,
	where $l_F$ and $l_T$ are the numbers of feature and target tokens, respectively (or simply the lengths of the image feature $F$ and the target embedding $T$). 
	Such complexity is prohibitive for a large $l_F$, hindering the existing methods from employing high-resolution image features in Transformers, which possess abundant fine-grained features~\cite{bosquet2020stdnet,sun2019deep,fcnn} that are beneficial for accurate 3D human shape and pose recovery.
	In this work, we propose two strategies to improve the Transformer framework to better exploit higher-resolution image features for high-quality 3D body shape and pose reconstruction.
	One of them is \emph{attention decoupling}.
	We notice that different from the original ViT~\cite{dosovitskiy2020image} where the image features are learned by attention operations, the 3D human Transformers~\cite{lin2021end,lin2021mesh} usually rely on Convolutional Neural Networks (CNNs) to extract these features, and the attention operations are mainly used to aggregate the image features to improve the target embeddings. 
	Thus, it is less important to model the feature-feature and feature-target correlations in Figure~\ref{fig: teaser}(a), as they do not have direct effects on the target. %
	We therefore propose to decouple the full attention operation into a target-feature attention and a target-target attention, which are cascaded together to avoid the quadratic complexity.
	As shown in Figure~\ref{fig: teaser}(b), the decoupled attention only has a computation and memory complexity of $\mathcal{O}(l_F l_T+l_T^2)$ which is linear with respect to the feature length $l_F$.

	The other strategy we propose for improving the previous 3D human Transformer is \emph{SMPL-based target representation.}
	Existing Transformers for 3D human shape and pose estimation mostly adopt a vertex-based target representation~\cite{lin2021end,lin2021mesh}, where the embedding length $l_T$ is often quite large, equaling the number of vertices on a body mesh. 
	Even with the proposed attention decoupling strategy, a large $l_T$ can still bring a considerable cost of computation and memory, which hinders the usage of high-resolution features.
	To address this issue, we introduce a new target representation based on the parametric body model SMPL~\cite{loper2015smpl}, with which we only need to learn a small number of target embeddings that account for the human body shape as well as 3D body part rotations.
	As illustrated in Figure~\ref{fig: teaser}(c), the SMPL-based representation $\mathcal{T}$ further lessens the computation and memory burden compared to the vertex-based representation with $l_\mathcal{T}  \ll l_T $.
	Combining the above two strategies leads to a much more concise attention learning process, which not only allows the utilization of high-resolution features in the Transformer, but also motivates us to explore more new designs for better 3D human shape and pose estimation.
	In particular, enabled by the lifted efficiency of our model, we introduce a multi-scale attention operation that effectively exploits the multi-scale information in a simple and  unified framework.
	Further, as the proposed target representation explicitly describes 3D relative rotations between body parts which are mostly local, we further propose a joint-aware attention module that emphasizes local regions around body joints to better infer the articulation status of the 3D human.

	To summarize, we make the following contributions: 
	\begin{itemize}
		\item[$\bullet$] By introducing the attention decoupling and SMPL-based target representation, we propose a new Transformer framework, SMPLer, that can exploit a large number of feature tokens for accurate and efficient 3D human shape and pose estimation. %

		\item[$\bullet$]  %
		Based on the two key designs above, we further develop multi-scale attention and joint-aware attention modules which significantly improve the reconstruction results. 

		\item[$\bullet$] Extensive experiments demonstrate that SMPLer performs favorably against the baseline methods  with better efficiency.
		In particular, compared to the state-of-the-art method~\cite{lin2021mesh}, the proposed algorithm lowers the MPJPE error by more than 10\% on Human3.6M~\cite{human3.6} with fewer than one-third of its parameters, showing the effectiveness of the proposed algorithm.
	\end{itemize}

	\section{Related Work}
	We first review the previous %
	methods for 3D human pose and shape estimation and then discuss recent advances in vision Transformers.
	
	\subsection{3D Human Shape and Pose Estimation}
	Recent years have witnessed significant progress in the field of monocular 3D human shape and pose estimation~\cite{bogo2016keep,tung2017self,pavlakos2018learning,lassner2017unite,omran2018neural,kolotouros2019convolutional,guler2019holopose,xu2019denserac,kolotouros2019spin,jiang2020coherent,zhang2020object,zeng20203d,zanfir2020weakly,song2020human,Choi_2020_ECCV_Pose2Mesh,georgakis2020hierarchical,Moon_2020_ECCV_I2L-MeshNet,zhang2020learning,Moon_2022_CVPRW_Hand4Whole,li2021hybrik,lin2021end,lin2021mesh,sengupta2021probabilistic,akhter2015pose,zanfir2021neural,joo2021exemplar,kolotouros2021probabilistic,dwivedi2021learning,sun2021monocular,zanfir2021thundr,zhang2021pymaf,kocabas2021spec,kocabas2021pare,kanazawa2019learning,arnab2019exploiting,sun2019human,doersch2019sim2real,kocabas2019vibe,luo20203d,choi2021beyond,lee2021uncertainty,wan2021encoder,xu20203d,xu2021_3d,moreno20173d,rong2021frankmocap,davydov2022adversarial,tiwari22posendf}.
	Due to the intrinsic ambiguity of 2D-to-3D mapping, this problem 
	is highly ill-posed and requires strong prior knowledge learned from large datasets to regularize the solutions.
	Indeed, the progress of 3D human shape and pose estimation largely relies on the development of powerful data-driven models.

	As a pioneer work, SMPLify~\cite{bogo2016keep} learns a Gaussian Mixture model from CMU marker data~\cite{cmu_mocap_data} to encourage plausible 3D poses, which however needs to conduct inference in a time-consuming optimization process.
	To address this low efficiency issue, more recent methods use deep learning models to directly regress the 3D human mesh in an end-to-end fashion, which have shown powerful capabilities in absorbing prior knowledge and discovering informative patterns from a large amount of data.
	Typically, GraphCMR~\cite{kolotouros2019convolutional} trains Graph Neural Networks (GNNs) for 3D human shape and pose estimation, which directly learns the vertex location on human mesh and is prone to outliers and large errors under challenging poses and cluttered backgrounds.
	In contrast, some other methods, such as SPIN~\cite{kolotouros2019spin} and RSC-Net~\cite{xu2021_3d}, circumvent this issue by training deep CNNs to estimate the underlying SMPL parameters, which have demonstrated robust performance for in-the-wild data, inspiring numerous new applications~\cite{wang2019re,xu2021texformer,rajasegaran2021tracking,li2021ai,liu2023sewformer}.

	Nevertheless, recent research in the machine learning community~\cite{dosovitskiy2020image,liu2021swin} reveals shortcomings of CNNs that they have difficulties in modeling long-range dependencies, which limits their capabilities for higher-quality representation learning.
	To overcome the above issue, METRO~\cite{lin2021end} and Mesh Graphormer~\cite{lin2021mesh} introduce Transformers to this task, significantly improving the reconstruction accuracy.
	However, these Transformer-based methods generally follow the architecture of ViT~\cite{dosovitskiy2020image} without considering the characteristics of 3D human shape and pose, and adopt a straightforward vertex representation, which hinders the network from exploiting high-resolution features for better performance.
	In contrast, we propose a decoupled attention formulation that is more suitable for accurate 3D human reconstruction with a large number of feature tokens.
	To better realize the concept of attention decoupling, we develop a new Transformer framework that can effectively exploit multi-scale (high and low) and multi-scope (global and local) information for better results.
	Moreover, we show that the parametric SMPL model can be combined with Transformers by introducing an SMPL-based target representation, which naturally addresses issues of vertex regression.

	\subsection{Transformers}
	With its remarkable capability of representation learning, the Transformer has become a dominant architecture in natural language processing~\cite{vaswani2017attention,devlin2018bert}.  
	Recently, it has been introduced in computer vision, and the applications include object detection~\cite{carion2020end}, image classification~\cite{dosovitskiy2020image}, image restoration~\cite{chen2020pre}, video processing~\cite{shi2022video}, and generative models~\cite{zhang2019self,jiang2021transgan}. 
	Most of these works employ an attention operation that has quadratic complexity regarding the feature length and thus is not suitable for our goal of exploiting higher-resolution features for more accurate 3D human shape and pose estimation.

	In this work, we demonstrate the effectiveness of a decoupled attention mechanism that has linear complexity with respect to the feature length. 
	We also introduce a compact target representation based on a parametric human model.
	Furthermore, we provide new insights in designing multi-scale and joint-aware attention operations for 3D human reconstruction, and the proposed network achieves better performance than the state-of-the-art Transformers with improved efficiency.

	\begin{figure*}[t]
		\centering
		\includegraphics[width=0.9\linewidth]{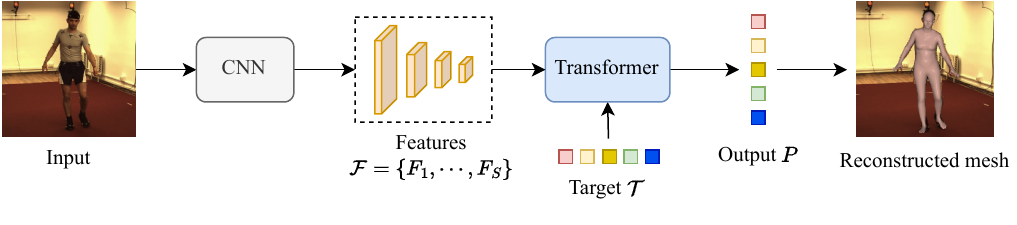}
		\vspace{-3mm}
		\caption{Overview of the proposed framework.
			Given a monocular input image, we first use a CNN backbone~\cite{sun2019deep} to extract image features $\mathcal{F}$, which are fed into the Transformer to reconstruct the 3D human body.
			The main ingredients of this framework are 1) an efficient decoupled attention module in the Transformer (Section~\ref{sec: efficient attention}), and 2) a compact target representation $\mathcal{T}$ based on parametric human model (Section~\ref{sec: compact representation}).
			More detailed descriptions of the Transformer architecture are provided in Figure~\ref{fig: hierarchical}.
		} 
		\label{fig: pipeline}
	\end{figure*}
	
		\begin{figure*}[t]
		\footnotesize
		\begin{center}
			\begin{tabular}{c}
				\hspace{0mm}
				\includegraphics[width = .98\linewidth]{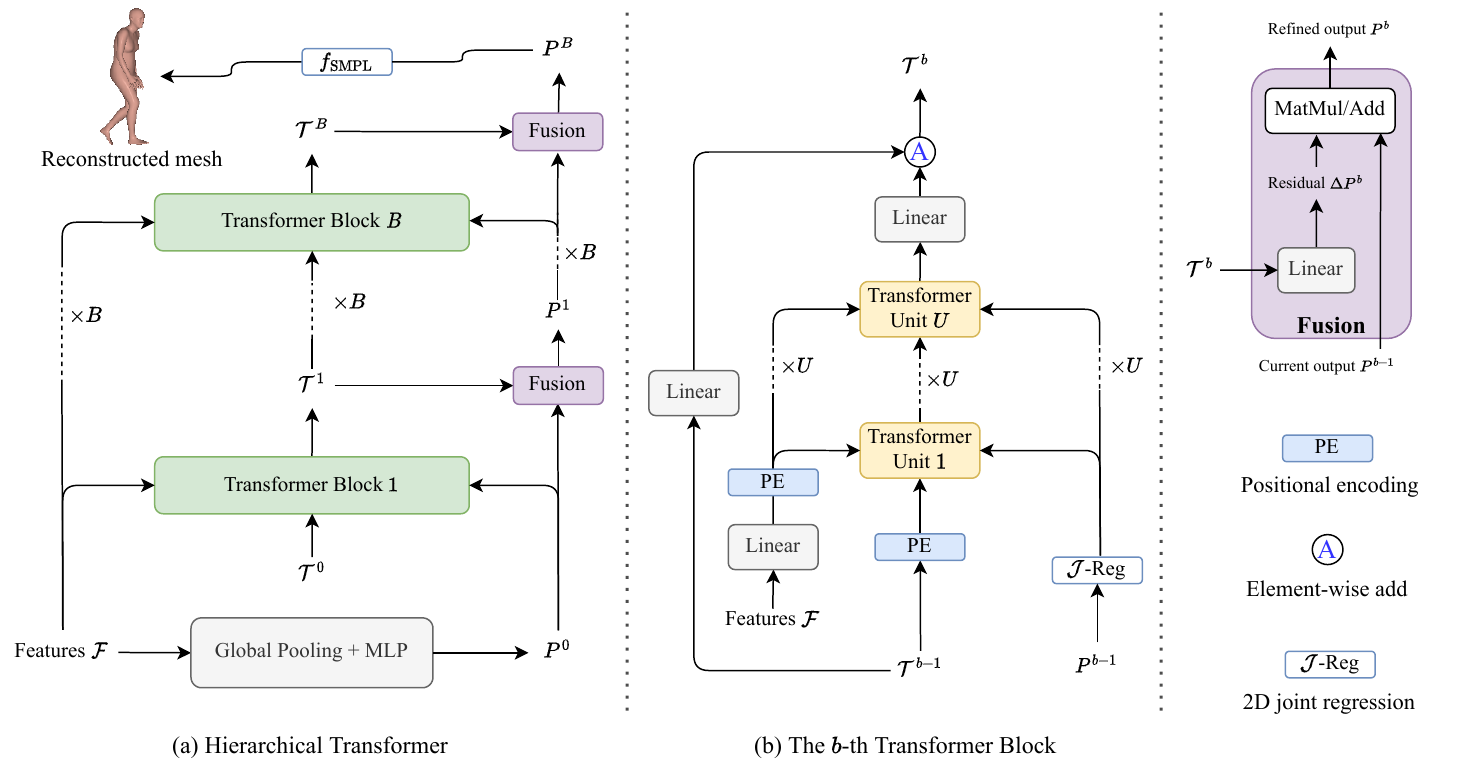} 
			\end{tabular}
		\end{center}
		\vspace{-3mm}
		\caption{Hierarchical architecture of our Transformer. 
			(a) shows an overview of the hierarchical architecture which corresponds to  the ``Transformer'' in Figure~\ref{fig: pipeline}.
			With the image features $\mathcal{F}$, we  progressively refine the initial estimation $P^0$ with $B$ Transformer Blocks (Eq.~\ref{eq: refine}). 
			In (b), each Transformer Block consists of $U$  Transformer Units, and each Unit is formulated as $h_\text{final}$ in Eq.~\ref{eq:h_final}. 
			The module ``$\mathcal{J}$-Reg'' represents 2D joint regression from the 3D estimation results, corresponding to Eq.~\ref{eq:smpl}-\ref{eq:2D_joints}.
		}
		\label{fig: hierarchical}
	\end{figure*}

	\section{Methodology}

	In this work, we propose a new 
	SMPL-based Transformer framework (SMPLer) for 3D human shape and pose estimation,
	which can exploit a large number of image feature tokens
	based on an efficient decoupled attention and a compact target representation.
	An overview is shown in Figure~\ref{fig: pipeline}.
	
	\subsection{Efficient Attention Formulation}\label{sec: efficient attention}
	The attention operation~\cite{vaswani2017attention} is the central component of Transformers, which can be formulated as:
	\begin{align}\label{eq: attn}
		h({Q},{K},{V}) = {f_\text{soft}} \left( \frac{({Q}  W_{q})  ({K}   W_{k})^{\top}}{\sqrt{d}} \right) ({V}   W_{v}),
	\end{align}
	where $f_\text{soft}$ is the softmax function along the row dimension. ${Q} \in \mathbb{R}^{l_Q \times d}$, and ${K}, {V} \in \mathbb{R}^{l_K \times d}$ are the input of this operation, representing query, key, and value, respectively.
	$l_Q$ is the length of the query ${Q}$, and $l_K$ is the length of ${K}$ and ${V}$.
	$d$ corresponds to the channel dimension.
	$W_{q},W_{k},W_{v} \in \mathbb{R}^{d \times d}$ represent learnable linear projection weights.
	We use a multi-head attention~\cite{vaswani2017attention} in our  implementation, where each head follows the formulation in Eq.~\ref{eq: attn}, and different heads employ different linear projection weights.
	Similar to prior work~\cite{vaswani2017attention},  we also use layer normalization~\cite{ba2016layer} and MLP in the attention operation, which are  omitted in Eq.~\ref{eq: attn} for brevity.

	Essentially, Eq.~\ref{eq: attn} models dependencies between pairs of tokens from ${Q}$ and ${K}$ with dot product. 
	The output $h(Q,K,V) \in \mathbb{R}^{l_Q \times d}$ can be seen as a new query embedding enhanced by aggregating information from ${V}$, and the aggregation weights are decided by the dependencies between ${Q}$ and ${K}$.
	Noticeably, when query, key and value are the same, Eq.~\ref{eq: attn} is called self-attention which we denote as $h_\text{self}(Q)=h(Q,Q,Q)$.
	When only key and value are the same while query is different, the operation becomes cross-attention, denoted as $h_\text{cross}(Q,K)=h(Q,K,K)$.
	\subsubsection{Full Attention}
	Existing Transformers for 3D human reconstruction~\cite{lin2021end,lin2021mesh} basically follow a ViT-style structure~\cite{dosovitskiy2020image}, which adopts the full attention formulation as shown in Figure~\ref{fig: teaser}(a).
	Mathematically, the full attention can be written as:
	\begin{align} \label{eq:full-attention}
		h_\text{self}(T \mathbin\Vert F),
	\end{align}
	where the self-attention $h_\text{self}(Q)$ is used, and the $Q$ is  a concatenation of the target embedding $T$ and the image feature $F$ along the token dimension, denoted by $T\mathbin\Vert F \in \mathbb{R}^{(l_T+l_F)\times d}$. 
	The target embedding represents the variable of interest, which corresponds to the class token in ViT and the  3D human representation in this work (see more explanations in Section~\ref{sec: compact representation}).
	
	As the image feature $F$ in Eq.~\ref{eq:full-attention} is involved in both query and key\footnote{This can be seen by replacing $Q,K,V$ in Eq.~\ref{eq: attn} with $T\mathbin\Vert F$.},  the full attention leads to a quadratic computation and memory cost with respect to the length of the image feature $l_F$, \ie $\mathcal{O}((l_F+l_T)^2)$,. %
	Consequently, previous works~\cite{lin2021end,lin2021mesh} only use low-resolution features in the Transformer where $l_F$ is small.
	In particular, suppose different resolution features from a CNN backbone are denoted as $\mathcal{F}=\{F_1, \cdots, F_S\}$ (see Figure~\ref{fig: pipeline}) where $S$ is the number of scales\footnote{The resolution decreases from scale $1$ to $S$.}, 
	Mesh Graphormer~\cite{lin2021mesh} only uses $F_S$ in the attention operation, and straightforwardly including higher-resolution features in Eq.~\ref{eq:full-attention}, \eg, $F_1$, would be computationally prohibitive.

	\subsubsection{Decoupled Attention}
	We notice that the attention operation serves different roles in ViT~\cite{dosovitskiy2020image} and the Transformers for 3D human shape and pose estimation~\cite{lin2021end,lin2021mesh}:
	the original ViT relies solely on attention to learn expressive image features, while the 3D human Transformer uses a deep CNN for feature extraction (Figure~\ref{fig: pipeline}), and the attention is mainly used to aggregate the image features for improving the target embeddings.
	Therefore, modeling feature-feature dependencies is less important here as it does not have a direct effect on the target, implying that it is possible to avoid the quadratic complexity by pruning the full attention.

	Motivated by this observation, we propose a decoupled attention that bypasses the feature-feature computations as shown in Figure~\ref{fig: teaser}(b). 
	It is composed of a target-feature cross-attention and a target-target self-attention, which can be written as:
	\begin{align} \label{eq:decoupled attention}
		h_\text{self}(h_\text{cross}(T, F)).
	\end{align}
	Notably, Eq.~\ref{eq:decoupled attention} has a computation and memory cost of $\mathcal{O}(l_F l_T + l_T^2)$, which is linear with respect to the feature length $l_F$.

	\subsection{Compact Target Representation} \label{sec: compact representation}
	While the attention decoupling strategy effectively relaxes the computation burden, a large $l_T$ may still hinder the utilization of high-resolution features in Eq.~\ref{eq:decoupled attention}. 
	Existing 3D human Transformers~\cite{lin2021end,lin2021mesh} usually recover the 3D body mesh by regressing the vertex locations $Y \in \mathbb{R}^{N \times 3}$, which leads to a redundant target representation $T \in \mathbb{R}^{N \times d}$, where the $i$-th row of $T$ is the embedding of the $i$-th vertex. 
	The length of this target embedding is often quite large ($N=6890$ by default for the SMPL mesh), resulting in heavy computation and memory cost in attention operations even after mesh downsampling.
	To address this issue, we devise a more compact target representation based on a parametric human body model SMPL.

	\subsubsection{Parametric Human Model} \label{sec: smpl}
	SMPL~\cite{loper2015smpl} is a flexible and expressive human body model that has been widely used for 3D human shape and pose modeling. 
	It is parameterized by a set of pose parameters $\theta \in \mathbb{R}^{H \times 3}$ and a compact shape vector $\beta \in \mathbb{R}^{1\times10}$.
	The body pose $\theta$ is defined by a skeleton rig with $H=24$ joints including the body root. 
	The $i$-th row of $\theta$ (denoted by $\theta_i$) is the rotation of the $i$-th body part in the axis-angle form whose skew symmetric matrix lies in the Lie algebra space $\mathfrak{so}$(3)~\cite{dai2015euler}.
	The body shape is represented by a low-dimensional space learned with principal component analysis~\cite{jolliffe2016principal} from a training set of 3D human scans, and $\beta$ is the coefficients of the principal basis vectors.
	
	With $\theta$ and $\beta$, we can obtain the 3D body mesh:
	\begin{align} \label{eq:smpl}
		Y \in \mathbb{R}^{N \times 3} = f_{\text{SMPL}}(\theta, \beta),
	\end{align}
	where $f_{\text{SMPL}}$ is the SMPL function~\cite{loper2015smpl} that gives the vertices $Y$ on a pre-defined triangle mesh.
	The 3D body joints $J$ can be obtained from the vertices via linear mapping using a pretrained  matrix $\mathcal{M} \in \mathbb{R}^{H \times N}$:
	\begin{align} \label{eq:3D_joints}
		J \in \mathbb{R}^{H \times 3} = \mathcal{M} {Y}.
	\end{align}

	With the 3D human joints, we can further obtain the 2D joints using weak perspective projection. 
	Denoting the camera parameters as $C \in \mathbb{R}^{3}$ that represents the scaling factor and the 2D principal-point translation in the projection process, the 2D joints can be obtained by:
	\begin{align} \label{eq:2D_joints}
		\mathcal{J} \in \mathbb{R}^{H \times 2} = \Pi_{C}({J}),
	\end{align}
	where $\Pi_{C}$ is the weak perspective projection function~\cite{hartley2003multiple}. 

	\subsubsection{SMPL-Based Target Representation} \label{sec: smpl_target_rep}
	Inspired by the compactness and expressiveness of SMPL, we replace the original vertex-based representation with an SMPL-based representation.
	Since the variables of interest are $\{\theta_i\}_{i=1}^H, \beta, C$ as introduced in Eq.~\ref{eq:smpl} and \ref{eq:2D_joints}, 
	we design the new target representation as
	$\mathcal{T} \in \mathbb{R}^{(H+2) \times d}$, where the first $H$ rows of $\mathcal{T}$  correspond to the $H$ body part rotations of $\theta$, and the remaining two rows describe the body shape $\beta$ and camera parameter $C$. 

	This new target representation conveys several advantages.
	First, the length of this representation is much shorter than the vertex-based one ($H+2 \ll N$), thereby facilitating the efficient design of our model as shown in Figure~\ref{fig: teaser}(c).
	Second, our target representation is able to restrict the solution to the SMPL body space, which naturally ensures smooth meshes, whereas the vertex-based representation is prone to outliers and may lead to spiky body surfaces as shown in Figure~\ref{fig: visual_comparison}.
	Third, the SMPL-based representation explicitly models the 3D rotation of body parts and thus is more readily usable in many applications, \eg, driving a virtual avatar. 
	In contrast, the vertices cannot be directly used  and have to be converted into rotations first (often in an iterative optimization manner), which is sub-optimal in terms of both efficiency and accuracy (see more details in Section~\ref{sec:compare_with_sota}).

	\subsection{Multi-Scale Attention} \label{sec: multi-scale}
	The above strategies, \ie attention decoupling and SMPL-based target representation, enable us to explore different resolution features in the Transformer framework, which inspires a multi-scale attention design for high-quality 3D human shape and pose estimation.

	\begin{figure}[t]
		\centering
		\includegraphics[width=0.65\linewidth]{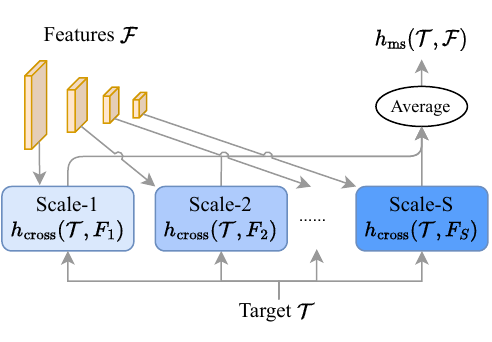}
		\vspace{-5mm}
		\caption{Jointly exploiting multi-scale features in the attention operation (see Eq.~\ref{eq:enhanced multi-scale attention} for more explanations).
		} 
		\label{fig: multi_scale_attn}
		\vspace{-3mm}
	\end{figure}

	\begin{figure}[t]
		\centering
		\includegraphics[width=0.99\linewidth]{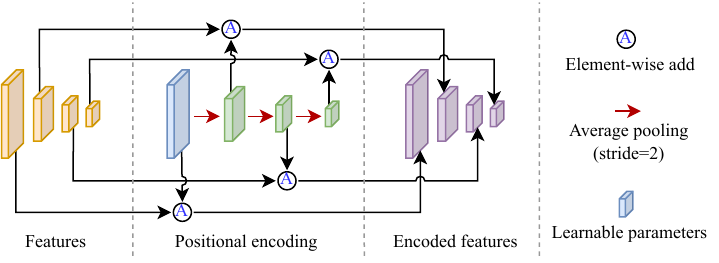}
		\vspace{-3mm}
		\caption{Pooling-based multi-scale positional encoding. We learn the positional encoding only for the highest-resolution feature, and the encodings for other scales are generated by average pooling, such that similar spatial locations across different scales have similar positional embeddings.
		} 
		\label{fig: pool_feat_pe}
		\vspace{-3mm}
	\end{figure}

	\subsubsection{Combining Multi-Scale Features}\label{sec:combine_multi_scale_features}
Our insight is that different resolution features are complementary to each other and should be collaboratively used for 3D human shape and pose estimation.
	A straightforward way to combine those features is to take each scale as a subset of tokens and concatenate all the tokens into a single feature embedding.
	Then the concatenated feature can be used as the $K$ of the cross-attention $h_\text{cross}(Q,K)$ in Eq.~\ref{eq:decoupled attention}. 
	The resulting multi-scale attention can be written as\footnote{We focus on improving the cross-attention part of the decoupled attention (Eq.~\ref{eq:decoupled attention}). The self-attention part is kept unchanged and  omitted  in the following sections for brevity.}:
	\begin{align} \label{eq:multi-scale}
		\tilde{h}_\text{ms}(\mathcal{T}, \mathcal{F}) =  h_\text{cross}(\mathcal{T},  F_1 \| F_2 \| \cdots \| F_S). 
	\end{align}
	However, this strategy uses the same projection weights for different scale features and thereby models target-feature dependencies in the same subspace~\cite{vaswani2017attention}, which ignores the characteristics of different scales and is less flexible for fully exploiting the abundant multi-scale information.
	
	To address this issue, we introduce an improved multi-scale strategy that  can be written as
	\begin{align} \label{eq:enhanced multi-scale attention}
		h_\text{ms} (\mathcal{T}, \mathcal{F}) = \frac{1}{S} \sum_{i=1}^{S} h_\text{cross}(\mathcal{T}, F_i),
	\end{align}
	where we employ different projection weights for each scale, and the output is an average of all scales as illustrated in Figure~\ref{fig: multi_scale_attn}.
	Whereas conceptually simple, Eq.~\ref{eq:enhanced multi-scale attention} effectively incorporates multi-scale image features into the attention computation, which differs sharply from existing works~\cite{lin2021end,lin2021mesh} that only rely on single-scale low-resolution features.

	\subsubsection{Multi-Scale Feature Positional Encoding} \label{sec:multi_scale_PE}

	As the attention operation itself in Eq.~\ref{eq: attn} is position agnostic, positional encoding is often used in Transformers to inject the position information of the input.
	Similar to ViT~\cite{dosovitskiy2020image}, we use the learnable positional encoding for the target and feature, which generally takes the form of $x+\phi$, where $\phi$ is a set of learnable parameters representing the position information of a token $x$. %

	In particular, the positional encoding of image features can be written as $\phi_i \in \mathbb{R}^{l_{F_i} \times d}$ ($i=1,2,\cdots, S$), which is directly added to feature $F_i$.
	Straightforwardly, we can learn positional encoding $\phi_i$ for all scales, which not only results in excessive parameters, but also ignores the relationship between the locations across different scales, \eg, similar spatial locations at different scales should have similar positional embeddings. 
	To address this issue, we adjust our strategy to only learn the positional embedding for the highest scale, \ie, $\phi_{1}$, and the embeddings for the other scales are produced by aggregating $\phi_{1}$: 
	\begin{align} \label{eq: pool_feat_pe}
		\phi_{i} = f_{\text{pool}}^{(2^{i-1})}(\phi_{1}),  
	\end{align}
	where $f_{\text{pool}}^{(2^{i-1})}$ is the average pooling with a stride and window size of $2^{i-1}$.
	In real implementation, we iteratively apply a stride-2 pooling layer to $\phi_1$ (see Figure~\ref{fig: pool_feat_pe}):
	\[\phi_{i} = f_{\text{pool}}^{(2)}(\phi_{i-1}), i=2,\cdots,S, \]  
	which is equivalent to Eq.~\ref{eq: pool_feat_pe} but requires slightly fewer computations.

	\begin{figure}[t]
		\centering
		\includegraphics[height=36mm]{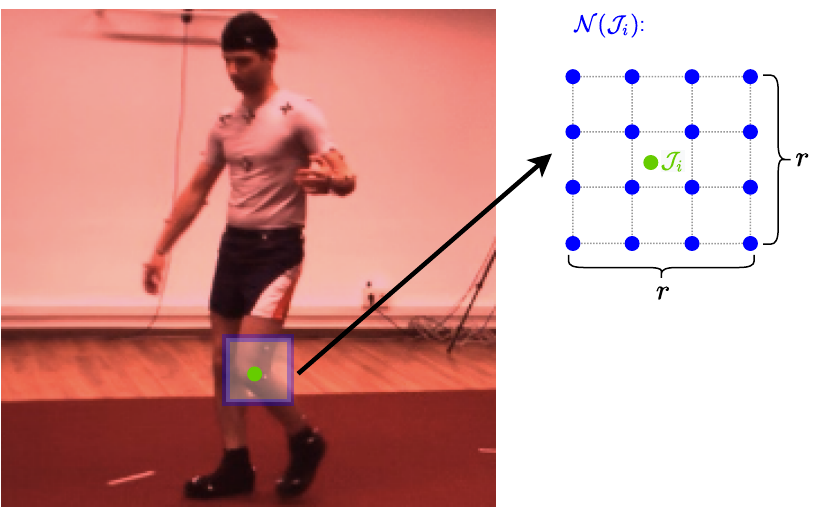}
		\caption{Illustration of the joint-aware attention that aggregates local features around human joints. See more details in Sec.~\ref{sec: joint-aware}.} 
		\label{fig: local_attn}
	\end{figure}
	
	\subsection{Joint-Aware Attention} \label{sec: joint-aware}
	In addition to the multi-scale approach, the SMPL-based target representation $\mathcal{T}$ also motivates the design of a joint-aware attention.
	Recall that the first $H$ rows of $\mathcal{T}$ describe the relative rotations of the $H$ body parts (Section~\ref{sec: smpl_target_rep}).
	As shown in Figure~\ref{fig: local_attn}, the local articulation status around human joints strongly implies the relative rotation between neighboring body parts.
	Thus, it can be beneficial to adequately focus on the local image features around joints in the attention operation to improve the first $H$ rows of $\mathcal{T}$ for better estimation of human pose.
	
	To this end, we devise a joint-aware attention operation, which modifies the cross attention $h_\text{cross}(Q,K)$ by restricting $K$ to a local neighborhood of the human joints.
	This operation can be written as:
	\begin{align}\label{eq: joint-aware attention}
		h_\text{ja}(\mathcal{T}_i, \mathcal{F}) = f_\text{soft} \left(\frac{({\mathcal{T}_i}W_q)  (F^{\mathcal{N}(\mathcal{J}_i)}_1 W_k)^\top}{\sqrt{d}}  + \eta \right) (F^{\mathcal{N}(\mathcal{J}_i)}_1 W_v),
	\end{align} 
	where $\mathcal{T}_i$ is the $i$-th row of $\mathcal{T}$, $i=1,\cdots, H$. 
	$\mathcal{N}(\mathcal{J}_i)$ denotes an image patch around the $i$-th joint $\mathcal{J}_i$ with size $r\times r$ as shown in Figure~\ref{fig: local_attn}, 
	and $F^{\mathcal{N}(\mathcal{J}_i)}_1  \in \mathbb{R}^{r^2 \times d}$ represents the local features sampled at $\mathcal{N}(\mathcal{J}_i)$ from $F_1$.
	Eq.~\ref{eq: joint-aware attention} has a similar form to Eq.~\ref{eq: attn}, where $\mathcal{T}_i$ and $F^{\mathcal{N}(\mathcal{J}_i)}_1$ serve as the $Q$ and $K$, respectively.
	It is noted that we only use the highest-resolution feature $F_1$ for the local attention here, as $\mathcal{N}(\mathcal{J}_i)$ can cover a large area on smaller feature maps such that the joint-aware attention on lower-resolution features becomes similar to the global attention in Eq.~\ref{eq:enhanced multi-scale attention}.
	Similar to the Swin Transformer~\cite{liu2021swin}, we incorporate a relative positional encoding $\eta \in \mathbb{R}^{1\times r^2}$ in the softmax function, which is bilinearly sampled from a learnable tensor  $\tilde{\eta} \in \mathbb{R}^{(r+1)\times (r+1)}$ according to the distance between $\mathcal{J}_i$ and pixels in $\mathcal{N}(\mathcal{J}_i)$.
	The local attention module has a computation and memory cost of $\mathcal{O}(Hr^2)$, which is almost negligible compared to the normal attention that attends to the image features globally.

\begin{figure}[t]
		\centering
		\includegraphics[width=0.93\linewidth]{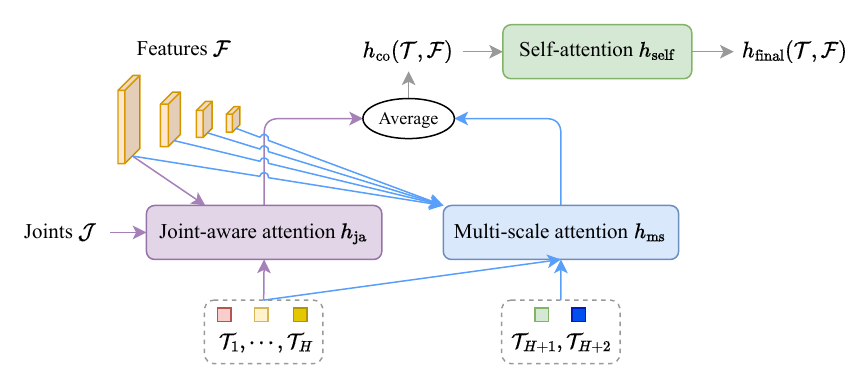}
		\vspace{-1mm}
		\caption{Combining the joint-aware and multi-scale attention. 
			Note that only the first $H$ rows of $\mathcal{T}$ are averaged in the ``Average'' operation (see Eq.~\ref{eq:combine attention} for more details). 
		} 
		\label{fig: combine_attn}
		\vspace{-3mm}
	\end{figure}

	Eventually, we combine the multi-scale attention (Eq.~\ref{eq:enhanced multi-scale attention}) and the joint-aware attention (Eq.~\ref{eq: joint-aware attention}) by simply taking the average as shown in Figure~\ref{fig: combine_attn}. 
	Denoting the combined attention as $h_\text{co}(\mathcal{T}, \mathcal{F}) \in \mathbb{R}^{(H+2)\times d}$, the $i$-th row of  the output  can be written as:
	\begin{align} \label{eq:combine attention}
		h_\text{co}(\mathcal{T}_i, \mathcal{F}) = \left\{ \begin{aligned}
			& \frac{1}{2} (h_\text{ja}(\mathcal{T}_i, \mathcal{F}) + h_\text{ms} (\mathcal{T}_i, \mathcal{F})),   ~i \le H \\
			& h_\text{ms} (\mathcal{T}_i, \mathcal{F}), ~i > H 
		\end{aligned}. \right.
	\end{align}
	Note that $h_\text{ja}(\mathcal{T}_i, \mathcal{F})$ is only defined for the $H$ body parts, \ie, $i=1,\cdots, H$, and thereby we directly use the multi-scale attention for the body shape $\beta$ and camera $C$ without averaging, which  correspond to $i=H+1, H+2$ in Eq.~\ref{eq:combine attention}.

	Following the attention decoupling in Eq.~\ref{eq:decoupled attention}, the final formulation of our attention module can  be written as 
	\begin{align}\label{eq:h_final}
	h_\text{final}(\mathcal{T}, \mathcal{F}) = h_\text{self}(h_\text{co}(\mathcal{T}, \mathcal{F})),
	\end{align}
	which is briefly illustrated in Figure~\ref{fig: combine_attn}.

	\begin{table*}[t]
		\footnotesize
		\centering
		\caption{Quantitative comparison against the state-of-the-art methods on Human3.6M and 3DPW datasets. 
			MPVE represents mean per-vertex error.
			Numbers in \textbf{bold} indicate the best in each column. %
		}
		\vspace{-1.5mm}
		\begin{tabular}{lccccccc}
			\toprule
			\multirow{2}{*}{Method} & \multirow{2}{*}{Parameters (M)}~~ & \multicolumn{2}{c}{Human3.6M} & & \multicolumn{3}{c}{3DPW}\\ 
			\cmidrule{3-4}\cmidrule{6-8}
			&& MPJPE $\downarrow$ & PA-MPJPE $\downarrow$ & & MPVE $\downarrow$ & MPJPE $\downarrow$ & PA-MPJPE $\downarrow$  \\
			\midrule
			HMR~\cite{kanazawa2018end} & $-$ & 88.0 & 56.8 && $-$ & $-$ & 81.3  \\
			GraphCMR~\cite{kolotouros2019convolutional} & 40.7 & $-$ & 50.1 && $-$ & $-$ & 70.2\\
			SPIN~\cite{kolotouros2019spin} & $-$ & $-$ & 41.1  && 116.4 & $-$ & 59.2\\
			RSC-Net~\cite{xu2021_3d} & 28.0 & 67.2 & 45.7 && 112.1 & 96.6 & 59.1 \\
			\revise{FrankMocap~\cite{rong2021frankmocap}} & $-$ & $-$ & $-$ && $-$ & 94.3 & 60.0 \\
			VIBE~\cite{kocabas2019vibe} & 42.7 & 65.6 & 41.4 && 99.1 & 82.9 & 51.9 \\
			Pose2Mesh~\cite{Choi_2020_ECCV_Pose2Mesh} & 72.8 & 64.9 & 47.0 && $-$ & 89.2 & 58.9 \\
			I2LMeshNet~\cite{Moon_2020_ECCV_I2L-MeshNet} & 135.7 & 55.7 & 41.1 && $-$ & 93.2 & 57.7\\
			\revise{PARE~\cite{kocabas2021pare}} & $-$ & $-$ & $-$ && 88.6 & 74.5 & 46.5 \\ 
			METRO~\cite{lin2021end}  & 231.8 & 54.0 & 36.7 && 88.2 & 77.1 & 47.9\\
			Mesh Graphormer~\cite{lin2021mesh}  & 215.7 & 51.2 & 34.5 && 87.7 & 74.7 & 45.6\\
			\midrule
			SMPLer & 35.6 & 47.0 & 32.8  && 84.7 & 75.7 & 45.2\\
			SMPLer-L & 70.2 & \bf 45.2 & \bf 32.4 & & \bf 82.0 & \bf 73.7 & \bf 43.4 \\
			\bottomrule
		\end{tabular}
		\vspace{-2mm}
		\label{tab:compare-h36m-3dpw}
	\end{table*}

	\subsection{Hierarchical Architecture}
	As shown in Eq.~\ref{eq: joint-aware attention}, an important issue of our current design is that the joint-aware attention relies on the 2D joints $\mathcal{J}$, which is supposed to be an output of our algorithm (see Eq.~\ref{eq:2D_joints}).
	In other words, we need $\mathcal{J}$ to reconstruct the 3D human and meanwhile need the 3D human to regress $\mathcal{J}$, which essentially leads to a chicken-and-egg problem.
	To circumvent this problem, we propose a hierarchical architecture (Figure~\ref{fig: hierarchical}) to iteratively refine the 2D joint estimation and 3D reconstruction results.
	
	We denote the output of the $b$-th stage in Figure~\ref{fig: hierarchical}(a) as $P^b=\{R_{\theta_1}^b, \cdots, R_{\theta_H}^b, \beta^b, C^b\}$, where $R_{\theta_i}^b \in \text{SO(3)}$ is the rotation matrix of the $i$-th body part, corresponding to $\theta_i$ (the $i$-th row of $\theta$).
	Then the refinement process can be written as:
	\begin{align} \label{eq: refine}
		\begin{aligned}
			\mathcal{T}^{b} &= f_{\text{TB}}^b (\mathcal{T}^{b-1}, {P}^{b-1}, \mathcal{F}), \\
			{P}^b &= f_\text{fusion}(\mathcal{T}^{b}, {P}^{b-1}), ~~~ b = 1,2,\cdots, B, 
		\end{aligned}
	\end{align}
	where $f_{\text{TB}}^b$ is the $b$-th Transformer Block, which improves the target embedding  $\mathcal{T}^{b-1}$ with image features $\mathcal{F}$ and the current 3D estimation $P^{b-1}$.
	$f_\text{fusion}$ is a fusion layer that generates the refined estimation $P^b$ based on the improved target embedding $\mathcal{T}^{b}$.
	As illustrated on the rightmost of Figure~\ref{fig: hierarchical}, in the fusion layer we first use a linear layer to map $\mathcal{T}^{b}$ into a set of residuals: 
	\begin{align}
		\Delta P^b = \{\Delta R_{\theta_1}^b, \cdots,  \Delta R_{\theta_H}^b, \Delta \beta^b, \Delta C^b\}, \nonumber
	\end{align}
	and then add the residuals to the current estimation $P^{b-1}$.
	Similar to \cite{zhou2019continuity}, we apply the Gram–Schmidt orthogonalization to the rotation residuals such that $\Delta R_{\theta_i}^b \in \text{SO}(3)$. We use matrix multiplication (MatMul in Figure~\ref{fig: hierarchical}) instead of addition to adjust the rotation parameters.
	
	As shown in Figure~\ref{fig: hierarchical}(a), the hierarchical architecture is composed of $B$ Transformer Blocks, and each Transformer Block is composed of $U$ Transformer Units in Figure~\ref{fig: hierarchical}(b).
	The Transformer Unit represents the aforementioned decoupled attention $h_\text{final}$ illustrated in Figure~\ref{fig: combine_attn}.
	To bootstrap this refinement process, we apply a global pooling layer and an MLP to the CNN features to obtain an initial coarse estimation $P^0$ similar to HMR~\cite{kanazawa2018end}.
	For the initial target embedding $\mathcal{T}^0$, a straightforward choice is to use a learned feature embedding as in DETR~\cite{carion2020end}.
	Nevertheless, we empirically find this choice makes the training less stable. Instead, we employ a heuristic strategy by combining the globally-pooled image feature and linearly-transformed $P^0$, \ie, $\mathcal{T}^0 = f_\text{global}(F_S) + f_\text{linear}(P^0)$, where $f_\text{global}$ is the global average pooling function.

	\subsection{Loss Function}\label{sec:loss}
	For training the proposed Transformer, we adopt the loss functions of Mesh Graphormer~\cite{lin2021mesh} which restrict the reconstructed human in terms of vertices and body joints:
	\begin{align}%
		\ell_\text{basic} = & w_Y \cdot \| Y - \hat{Y} \|_1 + w_J \cdot \|{J} - \hat{J} \|^2_2 + w_\mathcal{J} \cdot \| \mathcal{J} - \hat{ \mathcal{J}} \|^2_2, \nonumber
	\end{align}
	where $Y, J, \mathcal{J}$ are the predicted vertices, 3D joints, and 2D joints that are computed with Eq.~\ref{eq:smpl}-\ref{eq:2D_joints} using the final output of our Transformer, \ie, $P^B$ in Figure~\ref{fig: hierarchical}(a).
	$\hat{Y}, \hat{J}, \hat{\mathcal{J}}$ indicate the corresponding ground truth.
	$w_Y, w_J, w_{\mathcal{J}}$ are hyper-parameters for balancing different terms.
	Following~\cite{lin2021mesh}, we use $L_1$ loss for vertices and MSE loss for human joints.

	In addition, we include a rotation regularization term:
	\begin{align}\label{eq:rotation loss}
		\ell_\text{rotation} = w_R \cdot \frac{1}{H}\sum_{i=1}^{H}	\|R_{\theta_i} - \hat{R}_{\theta_i}\|_1,   
	\end{align}
	where we encourage the predicted rotation $R_{\theta_i} $  to be close to the ground-truth rotation matrix $\hat{R}_{\theta_i}$.
	$w_R$ is the weight of the loss.
	Eventually, our training loss is a combination of $\ell_\text{basic}$ and $ \ell_\text{rotation}$.
	We do not restrict the body shape $\beta$ as we empirically find no benefits from it.
	Note that $\ell_\text{rotation}$ can only be used in our Transformer which uses an SMPL-based target representation; it is in contrast to existing 3D human reconstruction Transformers that do not directly consider rotations.

		\begin{figure*}[t]
		\footnotesize
		\begin{center}
			\begin{tabular}{c}
				\includegraphics[width = 0.8\linewidth]{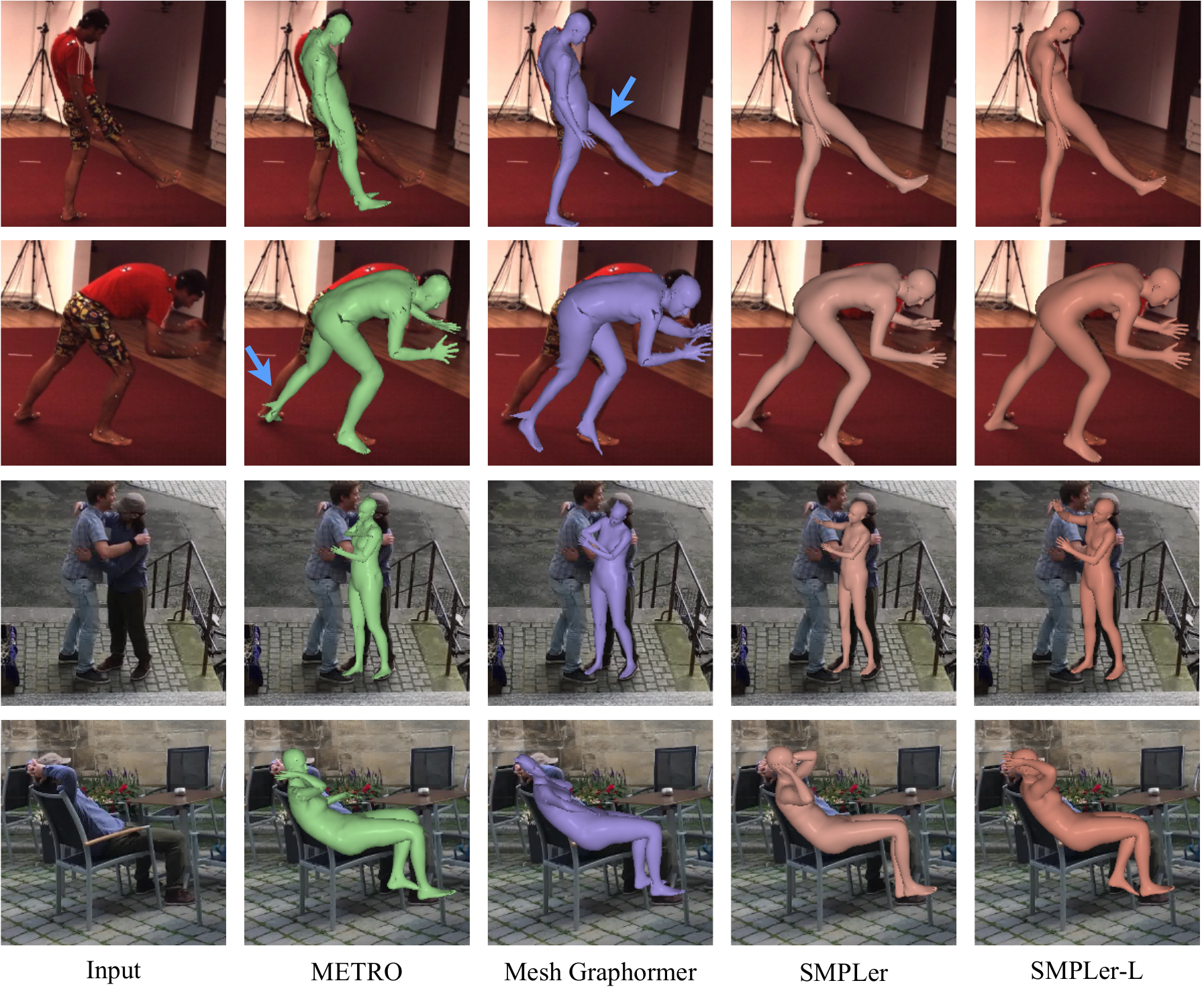} 
			\end{tabular}
		\end{center}
		\vspace{-2mm}
		\caption{Visual comparisons against the SOTA methods. The top two rows are from the Human3.6M dataset~\cite{human3.6}, and the bottom two rows are from the 3DPW dataset~\cite{3dpw}.
		}
		\vspace{-2mm}
		\label{fig: visual_comparison}
	\end{figure*}
	
{\Revise
	\subsection{Discussions}
	\minisection{Decoupled attention.}
	While a similar idea to our attention decoupling has been touched upon in~\cite{zhang2021temporal} for video classification, our work marks the pioneering exploration of it in monocular 3D human shape and pose estimation, addressing the intrincs limitations of the full attention formation of the existing works. 
	While we present the idea in a simplified manner in Figure~\ref{fig: teaser} for clarity, the innovation of our decoupled attention goes beyond a basic conceptual introduction. 
	Unlike the approach in \cite{zhang2021temporal} that relies solely on single-scale low-resolution features, we propose a multi-scale decoupled attention framework (Section~\ref{sec: multi-scale}). 
	This unique design allows the model to exploit both coarse and granular visual information, substantially improving the performance of 3D human shape and pose estimation.
	Notably, this multi-scale approach is enabled by the enhanced efficiency of our attention decoupling strategy, which allows incorporating higher-resolution features without prohibitive computational costs.
	
	In addition, we emphasize that effectively combining multi-scale features is not a straightforward endeavor, which requires dedicated algorithmic design and meticulous engineering efforts. 
	Particularly, instead of directly concatenating all scale features, our approach assigns different projection weights for each scale to model target-feature dependency in a scale-aware manner (Section~\ref{sec:combine_multi_scale_features}). 
	Moreover, we devise a pooling-based multi-scale positional encoding to better represent spatial information across varying scales, as detailed in Section~\ref{sec:multi_scale_PE}.

	\vspace{1mm}
	\minisection{SMPL-based target representation.}
	While SMPL has been used as output in prior methods~\cite{jiang2020coherent,kanazawa2018end,xu20203d,zhang2021pymaf}, our work marks the first time that the SMPL-based target representation is used in a Transformer framework.

	We emphasize the contribution of our SMPL-based target representation is rooted in its distinct advantages.
	First, it considerably reduces the computational cost as illustrated in Figure~\ref{fig: teaser}.
	Second, it ensures consistent, smooth SMPL mesh, avoiding the spiky outliers on human surfaces produced by vertex-based representations (Figure~\ref{fig: visual_comparison}).
	Third, it explicitly models body part rotations, facilitating its applications in driving virtual avatars (Figure~\ref{fig:virtual_avatar}).
	Furthermore, the SPML-based target representation allows the development of the joint-aware attention and motivates the hierarchical architecture of our Transformer, both of which are novel designs absent from previous works~\cite{jiang2020coherent,kanazawa2018end,xu20203d,zhang2021pymaf} and the concurrent work~\cite{zheng2023potter}, significantly improving the reconstruction results.
}
		\begin{figure*}[t]
		\footnotesize
		\begin{center}
			\begin{tabular}{c}
				\hspace{0mm}
				\includegraphics[width = .99\linewidth]{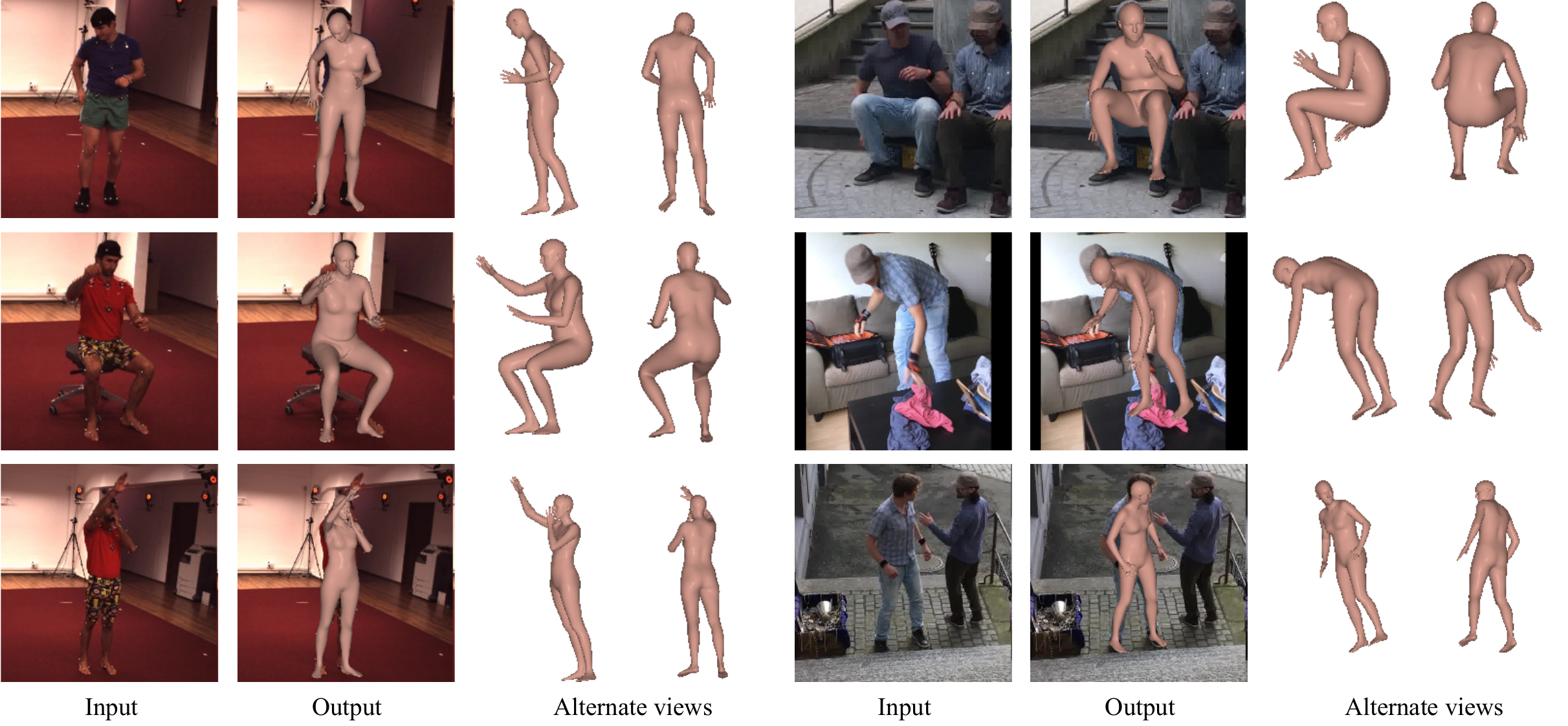} 
			\end{tabular}
		\end{center}
		\vspace{-2mm}
		\caption{Qualitative results of SMPLer. %
			For each example, the first column shows the input image, the second column shows the output in camera view, and the remaining columns show the predicted mesh from alternate viewpoints.
		}
		\vspace{-2mm}
		\label{fig: results}
	\end{figure*}
	
	\section{Experiments}
	
	\subsection{Implementation Details}
	For the network structure, we set the number of Transformer Blocks as $B=3$  and the number of Transformer Units in each block as $U=2$ by default.
	We use four attention heads for the Transformer  and set the feature sampling range of the joint-aware attention as $r=8$.
	\revise{It corresponds to a $32\times32$ region in the input image, which  is adequately large to encompass the vicinity of the joints.}
	We set the number of feature scales $S=4$.
	We use HRNet~\cite{sun2019deep} as the CNN backbone.
	We present two variants of the proposed Transformer: a base model SMPLer and a larger one SMPLer-L. 
	The two models use the same architecture, and the only difference is the channel dimension of the backbone, where we increase the channels by half for SMPLer-L.
	We use a batch size of 200 and train the model for 160 epochs.
	The loss weights in $\ell_\text{basic}$ and $ \ell_\text{rotation}$ are empirically set as $w_Y=100, w_J=1000, w_\mathcal{J}=100, w_R=50$.

	\vspace{1mm}
	\minisection{Dataset.}
	Similar to~\cite{lin2021end,lin2021mesh}, we extensively train our model by combining several human datasets, including Human3.6M~\cite{human3.6}, MuCo-3DHP~\cite{mehta2018single}, UP-3D~\cite{lassner2017unite}, COCO~\cite{coco}, MPII~\cite{mpii}. 
	Similar to~\cite{lin2021end,lin2021mesh}, we use the pseudo 3D mesh training data from \cite{Choi_2020_ECCV_Pose2Mesh,Moon_2020_ECCV_I2L-MeshNet} for Human3.6M.
	We follow the general setting where subjects S1, S5, S6, S7, and S8 are used for training, and subjects S9 and S11 for testing. We present all results using the P2 protocol~\cite{kanazawa2018end,kolotouros2019spin}. 
	We fine-tune the models with the 3DPW~\cite{3dpw} training data for 3DPW experiments.
	
	\vspace{1mm}
	{\Revise
	\minisection{Metrics.}
	We mainly use mean per-joint position error (MPJPE) and Procrustes-aligned mean per-joint position error (PA-MPJPE) for evaluation.
	MPJPE is defined as:
	\begin{align} \label{eq: mpjpe}
			\frac{1}{H} \sum_{i=1}^{H} \|J_i - \hat{J}_i \|_2,
		\end{align}
	where $J_i$ is the $i$-th row of $J$, \ie, the coordinate of the $i$-th human joint,
	and $\hat{J}$ represents the ground truth.
	Eq.~\ref{eq: mpjpe} directly measures the joint-to-joint error, which can be influenced by global factors, including scaling, global rotation and translation. 
	PA-MPJPE  is defined as:
	\begin{align} \label{eq: pa-mpjpe}
			& \min_{s,R,t} \frac{1}{H} \sum_{i=1}^{H} \|sJ_i R+t - \hat{J}_i \|_2, \\
			& \text{s.t.}  ~~ s\in \mathbb{R},  R \in \text{SO(3)}, t \in \mathbb{R}^{1\times 3}, \nonumber
		\end{align}
	where $s, R, t$ represent scaling, global rotation and translation, respectively.
	Instead of directly computing the joint-to-joint error, Eq.~\ref{eq: pa-mpjpe} first aligns the prediction $J$ to the ground truth $\hat{J}$ with a scaled rigid transformation (a closed-form solution for the alignment is given by Procrustes analysis~\cite{schonemann1966generalized}).
	Thus, PA-MPJPE is able to exclude the global factors and focus on the intrinsic human body shape and pose, emphasizing the measurement of relative position between adjacent body parts. }

		\begin{figure}[t]
		\centering
		\includegraphics[width=0.5\linewidth]{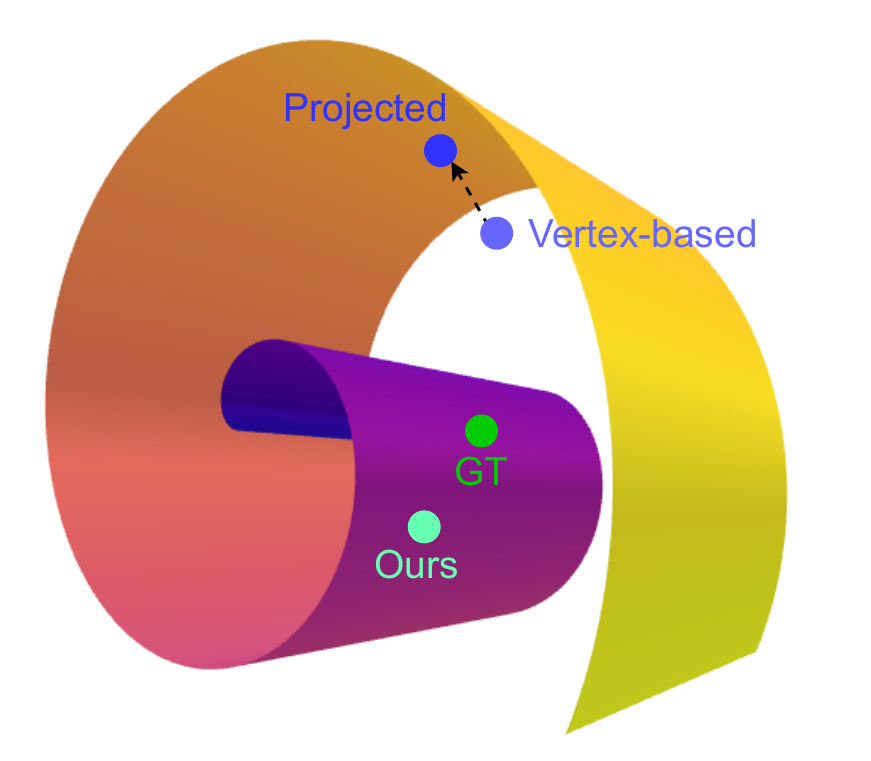}
		\vspace{-2mm}
		\caption{Illustrating the manifold of the desired SMPL human meshes.
			The proposed target representation guarantees our method always moves along this manifold, while the existing Transformers may produce results off the desired space, resulting in less accurate estimation and inconvenience in many applications, \eg, controlling virtual avatars. 
			As a straightforward remedy for avatar control, one can project the undesired  results to the SMPL manifold with iterative optimization, which however is often time-consuming and prone to accumulative errors.
			Note  that the real SMPL manifold is embedded in a 20670-dimensional space (6890 3D vertices), and here we simply show the concept in 3D for ease of visualization.
		} 
		\label{fig: manifold}
		\vspace{-3mm}
	\end{figure}

	\subsection{Comparison with the State of the Arts} \label{sec:compare_with_sota}
	\minisection{Quantitative evaluation.}
	We compare against the state-of-the-art 3D human shape and pose estimation methods, including CNN-based~\cite{kanazawa2018end,kolotouros2019spin,xu2021_3d,Moon_2020_ECCV_I2L-MeshNet,kocabas2019vibe,kocabas2021pare,rong2021frankmocap}, GNN-based~\cite{kolotouros2019convolutional,Choi_2020_ECCV_Pose2Mesh}), and Transformer-based~\cite{lin2021end,lin2021mesh}. 
	As shown in Table~\ref{tab:compare-h36m-3dpw}, the proposed Transformer performs favorably against the baseline approaches on both the Human3.6M and 3DPW datasets.
	Notably, compared to Mesh Graphormer~\cite{lin2021mesh}, the proposed SMPLer and SMPLer-L lowers the MPJPE error on Human3.6M by 8.2\% and 11.7\% respectively, while using only 16.5\% and 32.5\% of its parameters,  clearly demonstrating the effectiveness of our algorithm.
	
	\vspace{1mm}
	\minisection{Qualitative evaluation.}
	For a more intuitive understanding of the results, we also provide visual comparisons in Figure~\ref{fig: visual_comparison}. 
	As a typical example, in the first row of Figure~\ref{fig: visual_comparison}, existing approaches cannot well pose the human legs as they only use low-resolution features in the Transformer and thus are prone to errors under self-occlusions and challenging poses.
	In contrast, the proposed SMPLer is able to better exploit the abundant image features in a multi-scale (coarse and fine) and multi-scope (global and local) manner, which effectively improves the estimation performance in complex scenarios.
	On the other hand, since existing Transformers mostly rely on a vertex-based target representation, their results do not always lie on the SMPL mesh manifold as illustrated in Figure~\ref{fig: manifold}. 
	This leads to spiky outliers on the human surface as shown in the second row of Figure~\ref{fig: visual_comparison} or even distorted meshes in the fourth row of Figure~\ref{fig: visual_comparison}.
	By contrast, our method introduces an SMPL-based target representation, which naturally guarantees the solutions to lie on the smooth human mesh space (Figure~\ref{fig: manifold}) and thereby achieves higher-quality results  as shown in Figure~\ref{fig: visual_comparison}.

	In addition, we show more qualitative results of SMPLer with alternative views in Figure~\ref{fig: results}, where the proposed network performs robustly under diverse poses and complex backgrounds.
	In particular, the results show that our model is able to accurately recover global rotation relative to the camera coordinate frame, which is also validated by the improvements over MPJPE and MPVE metrics.

\begin{figure}[t]
		\centering
		\includegraphics[width=0.99\linewidth]{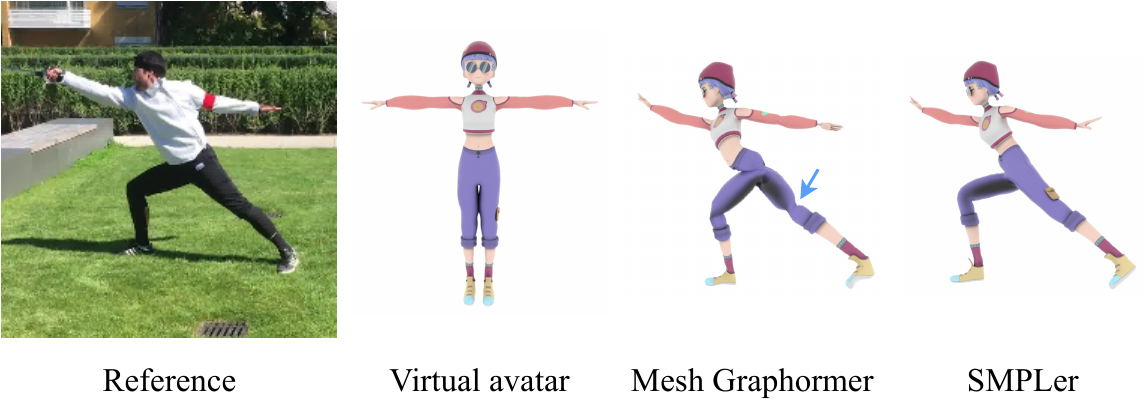}
		\vspace{-3mm}
		\caption{Thanks to the SMPL-based target representation, the proposed method can be well applied to control virtual avatars, while the existing vertex-based Transformers are error-prone. The virtual character is from Mixamo~\cite{mixamo}.
		} 
		\label{fig:virtual_avatar}
		\vspace{-2mm}
	\end{figure}

	\vspace{1mm}
	\minisection{Controlling virtual avatars.}
	An important application of 3D human shape and pose estimation is to control virtual avatars, \eg, in Metaverse.
	As our SMPL-based target representation explicitly models the 3D rotations of body parts, the proposed SMPLer can be easily used to drive virtual humans. An example is shown in Figure~\ref{fig:virtual_avatar}.
	In contrast, the previous Transformers take vertices as the output which are not directly applicable here and have to be converted into rotations first for this task.
	More specifically, the predicted vertices need to be fitted to the SMPL model (Eq.~\ref{eq:smpl}) in an iterative optimization manner, which corresponds to projecting the results onto the SMPL manifold in Figure~\ref{fig: manifold}.
	Compared to the proposed solution which is essentially one-step, the two-step approach (vertices $\rightarrow$ rotations) not only leads to inefficiency issues due to the time-consuming fitting process, but also suffers from low accuracy caused by accumulative errors.
	
	To quantitatively evaluate the rotation accuracy, we introduce a new metric, called mean per-body-part rotation error (MPRE), which is defined as:
	\begin{align} \label{eq:mpre}
		\text{MPRE} = \frac{180}{\pi H} \sum_{i=1}^{H}	\arccos(\frac{\text{trace}(R_{\theta_i} \hat{R}_{\theta_i}^\top) - 1}{2}),
	\end{align}
	where $R\hat{R}^\top$ represents the rotation matrix between the predicted rotation $R$ and the ground truth $\hat{R}$ (this can be seen from $R = (R\hat{R}^\top) \hat{R}$).
	Essentially, Eq.~\ref{eq:mpre} describes the rotation angle $\omega$ (in degree) between the prediction and ground truth, as the three eigenvalues of a rotation matrix are 1 and $e^{\pm i\omega}$, and thus  $\text{trace}(R\hat{R}^\top)=1+e^{i\omega}+e^{ -i\omega}=1+2\cos(\omega)$. 
	
	As the ground-truth rotations of Human3.6M are not available, we evaluate the rotation accuracy on 3DPW. 
	The proposed SMPLer achieves an MPRE of 9.9, significantly outperforming the 57.0 of Mesh Graphormer, which demonstrates the advantage of the proposed method in controlling virtual avatars.
	The visual comparison in Figure~\ref{fig:virtual_avatar} further verifies the improvement.

	\subsection{Analysis and Discussions}
	We conduct a comprehensive ablation study on Human3.6M to investigate the capability of our method.
	We also provide more analysis and discussions about the model efficiency as well as the attention mechanism.
	We use SMPLer as our default model in the following sections.

	\minisection{Decoupled attention and SMPL-based representation.}
	We propose a new Transformer framework that is able to exploit high-resolution features for accurate 3D human reconstruction.
	At the core of this framework are a decoupled attention module and an SMPL-based target representation as introduced in Section~\ref{sec: efficient attention} and \ref{sec: compact representation}.
	To analyze the effectiveness of these designs, we study different choices of the attention operation (full \vs decoupled) and the target representation (vertex-based \vs SMPL-based) in Table~\ref{tab:full-decouple}.

	As shown by Table~\ref{tab:full-decouple}(b), the straightforward approach of using a full attention and a vertex-based representation leads to prohibitive memory and computation cost for exploiting multi-resolution image features $\mathcal{F}$.
	With limited computational resources, we cannot train this model with a proper batch size.
	While the decoupled attention can well reduce the model complexity (Table~\ref{tab:full-decouple}(c)), the memory footprint is still  large due to the high dimensionality of the vertex-based target representation.
	In contrast, the proposed Transformer combines the decoupled attention and SMPL-based representation (Table~\ref{tab:full-decouple}(d)), which significantly lessens the memory and computational burden, and thereby achieves more effective utilization of multi-scale features.
	Note that the feature dimensionality is not a computational bottleneck for our network anymore, as it only leads to a marginal computation overhead for employing high-resolution features as shown by Table~\ref{tab:full-decouple}(d) and (e).

	\vspace{1mm}
	\minisection{Effectiveness of the multi-scale attention.}
	As introduced in Section~\ref{sec: multi-scale}, we propose an attention operation $h_\text{ms}$ for better exploiting multi-scale image information.
	As shown in Table~\ref{tab:ablation-multiscale}, using single-scale feature for 3D human shape and pose estimation, either the low-resolution (only scale $S$) or the high-resolution one (only scale 1), leads to inferior results compared to our full model.

	On the other hand, unifying multi-scale features in Transformer is not a trivial task. 
	As introduced in Section~\ref{sec: multi-scale}, instead of simply concatenating different resolution features into a single feature vector ($\tilde{h}_\text{ms}$ in Eq.~\ref{eq:multi-scale}), we separately process each scale with different projection weights, \ie, $h_\text{ms}$ in Eq.~\ref{eq:enhanced multi-scale attention}.
	As shown in Table~\ref{tab:ablation-multiscale}, the straightforward concatenation method cannot fully exploit the multi-scale information due to the undesirable restriction of using the same projection weights for different scales, resulting in less accurate 3D human estimation compared to the proposed approach.
	
	In addition, we apply multi-scale feature positional encoding  to supplement position information for the attention operation.
	As shown in Table~\ref{tab:ablation-ftpe}, the model without feature positional encoding (w/o FPE) suffers from a major performance drop, especially for MPJPE. 
	Furthermore, instead of directly learning the feature positional embedding for all scales (all-scale FPE), we propose a pooling-based strategy Eq.~\ref{eq: pool_feat_pe} as illustrated in Figure~\ref{fig: pool_feat_pe} to better handle the spatial relation across scales.
	Compared to ``all-scale FPE'' in Table~\ref{tab:ablation-ftpe}, this strategy further improves the  3D human shape and pose estimation results.

	\begin{table*}[t]
		\caption{Comparison across different choices of the attention operation and the target representation. 
			The GPU memory and GFlops are measured for a batch size of 1. 
			Due to the immense GPU memory requirement, we are not able to train (b) and (c) with a proper batch size.
			(d) corresponds to our default setting of SMPLer.
		}
		\centering
		\footnotesize
		\label{tab:full-decouple}
		\begin{tabular}{lccc|cccc}
			\toprule
			& Attention & Target  & Feature   & Memory (G)   & GFlops & MPJPE & PA-MPJPE \\
			\midrule
			(a) & Full & Vertex-based & $F_S$ & 0.49 & 8.1 & 51.9 & 36.0 \\
			(b) & Full & Vertex-based & $\{F_1,\cdots, F_S\}$ & 7.3 & 26.0 & $-$ & $-$ \\
			(c) & Decoupled & Vertex-based & $\{F_1,\cdots, F_S\}$ & 1.15 & 11.0 & $-$ & $-$ \\
			(d) & Decoupled & SMPL-based & $\{F_1,\cdots, F_S\}$ & 0.44 & 8.7 & 47.0 & 32.8 \\
			(e) & Decoupled & SMPL-based &  $F_S$ & 0.37 & 7.8 & 48.3 & 34.0 \\
			\bottomrule     
		\end{tabular}
		\vspace{-3mm}
	\end{table*}
	
	\begin{table}[t]
		\centering
		\caption{Effectiveness of the multi-scale attention.}
		\label{tab:ablation-multiscale}
		\begin{tabular}{lcc}
			\toprule
			Method           & ~MPJPE~ & ~PA-MPJPE~ \\
			\midrule
			only scale $S$           &   48.3    &  34.0        \\
			only scale $1$           &  49.5     &   33.7      \\
			concatenation & 48.5 & 33.9 \\
			full model        &   47.0    &    32.8    \\
			\bottomrule
		\end{tabular}
	\end{table}
	
	\begin{table}[t]
		\centering
		\caption{Effectiveness of the feature positional encoding (FPE).}
		\label{tab:ablation-ftpe}
		\begin{tabular}{lcc}
			\toprule
			Method           & ~MPJPE~ & ~PA-MPJPE~ \\
			\midrule
			w/o FPE & 50.4 & 33.7 \\
			all-scale FPE & 49.0 & 33.6 \\
			full model     &   47.0    &    32.8    \\
			\bottomrule
		\end{tabular}
	\end{table}
	
	\begin{table}[t]
		\centering
		\caption{Effectiveness of the joint-aware attention.}
		\label{tab:ablation-joint-aware}
		\begin{tabular}{lcc}
			\toprule
			Method           & ~MPJPE~ & ~PA-MPJPE~ \\
			\midrule
			w/o $h_\text{ja}$          &   51.4    &  34.5        \\
			w/o $\eta$  & 48.7 & 33.3 \\
			full model       &   47.0    &    32.8     \\
			\bottomrule
		\end{tabular}
	\end{table}
	
	\begin{table}[t]
		\centering
		\caption{Effect of different number of  Transformer Blocks and Units in the hierarchical architecture. $B=3, U=2$ is the default setting.}
		\label{tab:ablation-hierarchical}
		\begin{tabular}{lcc}
			\toprule
			Method           & ~MPJPE~ & ~PA-MPJPE~ \\
			\midrule
			$B=1, U=2$ & 48.6 & 34.4 \\
			$B=2, U=2$ & 47.4 & 33.1 \\
			$B=3, U=2$      &   47.0    &    32.8     \\
			$B=3, U=1$         &   48.5    &  33.9        \\
			$B=3, U=3$         &   47.1    &  32.8   \\
			\bottomrule
		\end{tabular}
	\end{table}
	
	\begin{table}[t]
		\centering
		\caption{Running speed of different methods measured on the same computer with an Intel Xeon E5-2620 v3 CPU and an NVIDIA Tesla M40 GPU. 
			The GFlops are calculated for a single input image with size 224$\times$224.}
		\label{tab:runtime}
		\begin{tabular}{lcc}
			\toprule
			Method   & Speed (fps) $\uparrow$ & GFLops $\downarrow$ \\
			\midrule
			METRO          &   33.5  & 50.5   \\
			Mesh Graphormer   & 34.6 & 45.4 \\
			SMPLer   &  \bf 96.0&\bf 8.7  \\
			SMPLer-L  &  \underline{73.9} & \underline{16.5}  \\
			\bottomrule
		\end{tabular}
	\end{table}

	\vspace{1mm}
	\minisection{Effectiveness of the joint-aware attention.}
	Motivated by the SMPL-based target representation, we propose a joint-aware attention $h_\text{ja}$ in Eq.~\ref{eq: joint-aware attention} to better exploit local features around human joints.
	As shown in Table~\ref{tab:ablation-joint-aware}, the model without $h_\text{ja}$ suffers from a significant performance drop, showing the importance of this module in inferring relative body part rotations.
	Further, we include a relative positional encoding in Eq.~\ref{eq: joint-aware attention} to model the spatial relation between target embedding and image features, which slightly improves the performance as  shown in Table~\ref{tab:ablation-joint-aware} (``w/o $\eta$'').

\begin{figure}[t]
		\centering
		\includegraphics[width=0.99\linewidth]{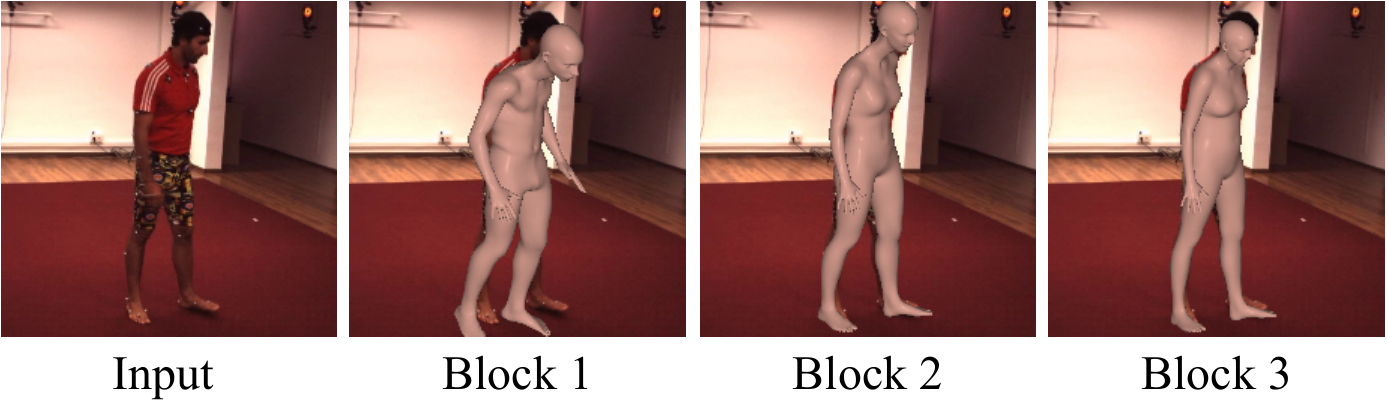}
		\vspace{-3mm}
		\caption{Output of Block 1, 2, and 3 of SMPLer. The reconstruction result is progressively refined in the hierarchical architecture.
		} 
		\label{fig: refine_process}
	\end{figure}

	\vspace{1mm}
	\minisection{Analysis of the hierarchical architecture.}
	We apply a hierarchical architecture in Figure~\ref{fig: hierarchical} that is composed of  multiple Transformer Blocks and multiple Transformer Units to
	progressively refine the 3D human estimation results. 
	We visualize the refinement process in Figure~\ref{fig: refine_process}, where the reconstruction becomes more accurate after more Blocks.

	Moreover, we investigate the effect of different settings of the hierarchical architecture. 
	As shown in Table~\ref{tab:ablation-hierarchical}, the performance gain of adding more Blocks and Units is significant at the start but converges quickly after  $B>1$ and $U>1$. 
	Therefore, we refrain from adding even more blocks and take $B=3, U=2$ as the default setting for a better tradeoff
	between performance and computation cost.

	\begin{figure}[t]
		\centering
		\includegraphics[width=0.99\linewidth]{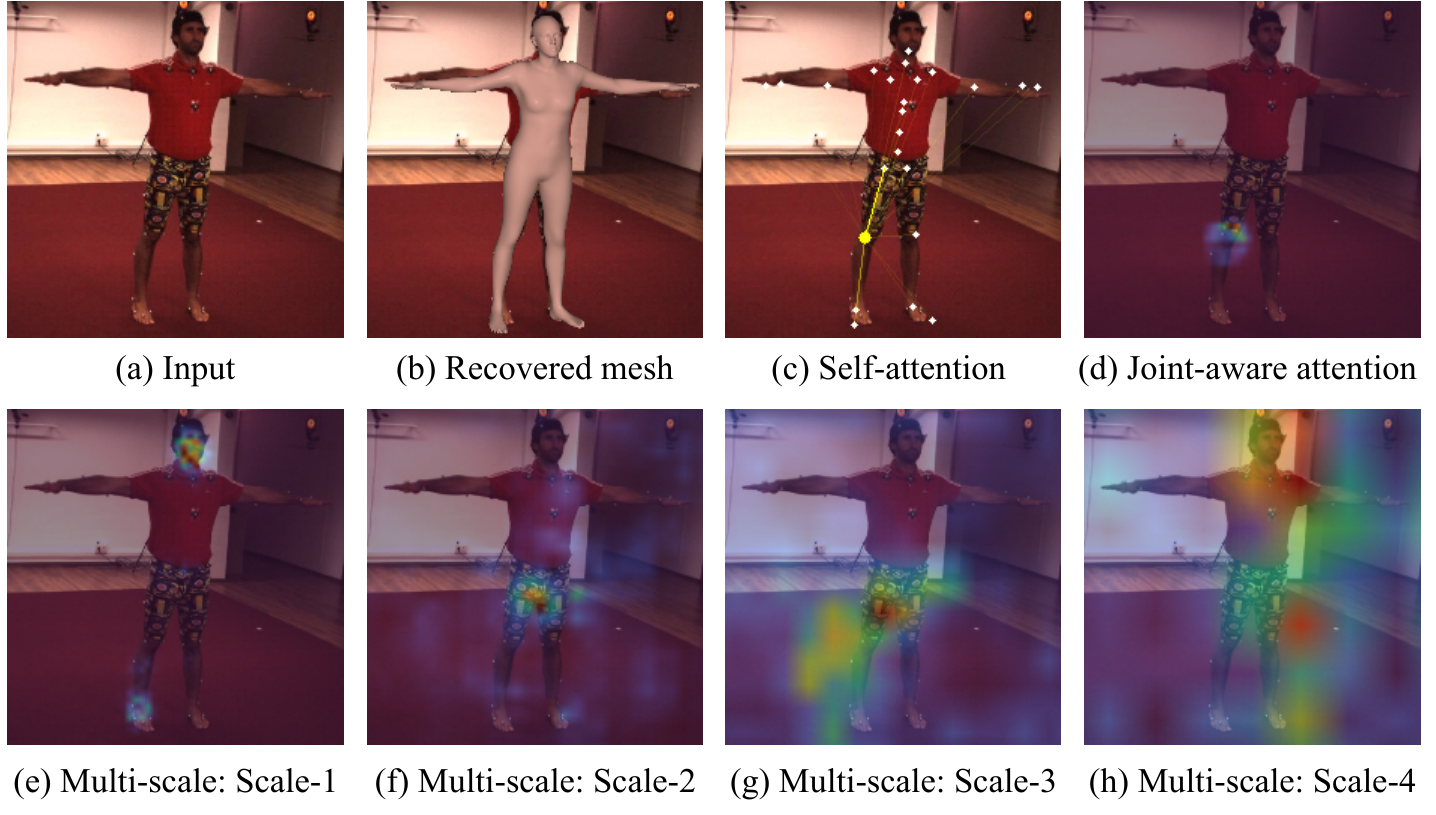}
		\vspace{-3mm}
		\caption{Visualization of the attention learned by SMPLer. The query joint is the right knee (the yellow point in (c)). For the self-attention in (c), brighter color indicates stronger
			interactions. For (d)-(h), redder color indicates larger attention response.
		} 
		\label{fig: vis_attn}
		\vspace{-3mm}
	\end{figure}
	
	\begin{figure}[t]
		\centering
		\includegraphics[width=0.7\linewidth]{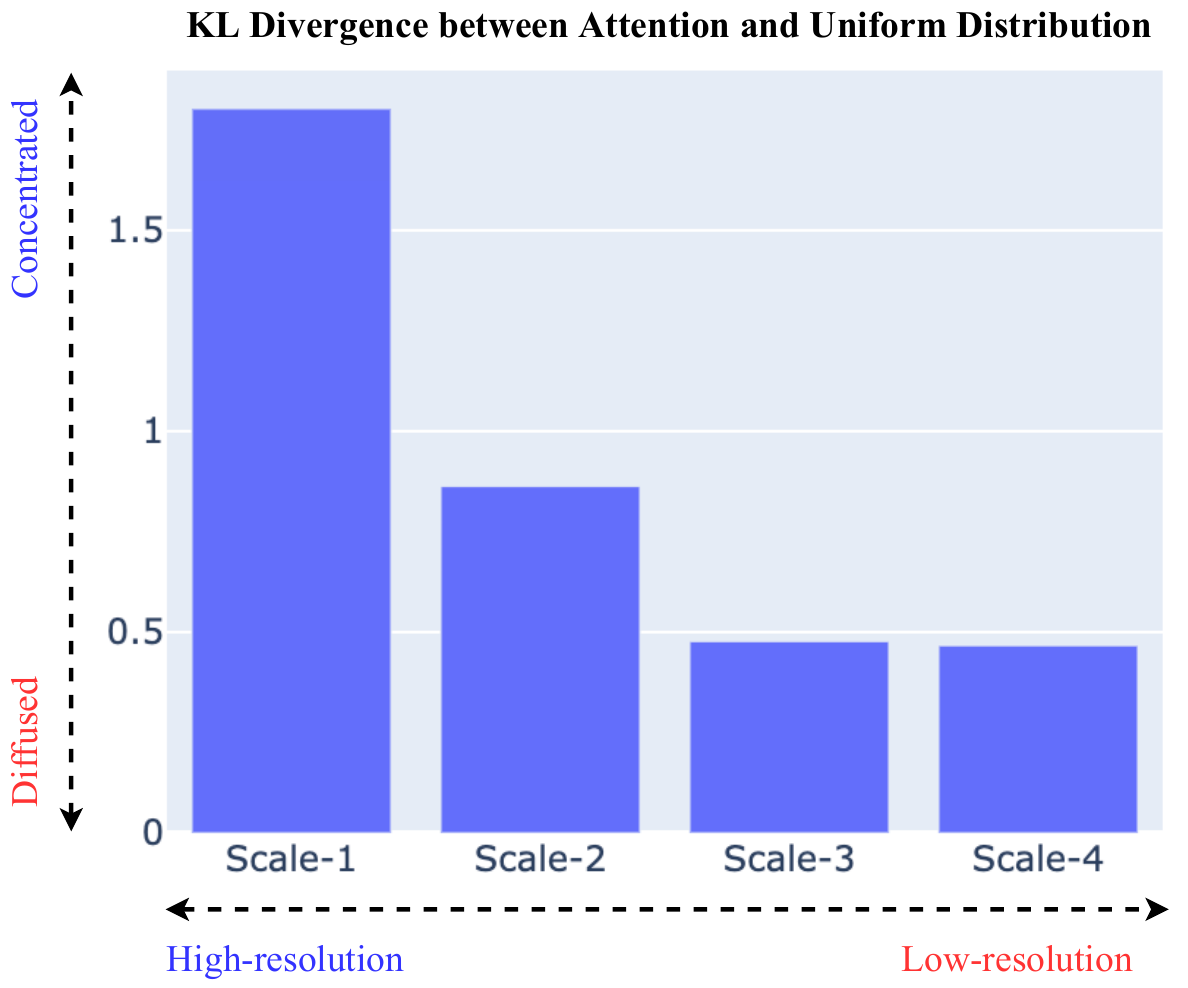}
		\vspace{-3mm}
		\caption{Attention distributions of different scales averaged on Human3.6M. We use the KL divergence between the attention map and a uniform distribution to quantitatively measure the spatial diffuseness of the attention. A higher divergence from the uniform distribution indicates lower diffuseness in space; in other words, the attention in that scale is more concentrated at fine-grained features.
		} 
		\label{fig: attn_stats}
		\vspace{-3mm}
	\end{figure}

	\vspace{1mm}
	\minisection{Attention visualization.}
	For a more comprehensive study of the proposed attention modules, we visualize the learned attention maps in Figure~\ref{fig: vis_attn}. 
	First, we show the multi-scale attention $h_\text{ms}$ in Figure~\ref{fig: vis_attn}(e)-(h), where the different scales correspond to the cross attention modules in Figure~\ref{fig: multi_scale_attn}.
	The attention maps for higher resolutions, \eg, in Figure~\ref{fig: vis_attn}(e), tend to be more concentrated on fine-grained features of specific body parts.
	In particular, the proposed Transformer relies on the foot and head poses to help infer body part rotation around the knee. 
	By contrast, for lower resolutions, \eg, in Figure~\ref{fig: vis_attn}(h), the attention gets more diffused in a wider space, indicating that the multi-scale attention can exploit global information, including the background, to decide the global rotation and scale of the body.
	
	To further analyze the attention of different scales, we also provide quantitative results in Figure~\ref{fig: attn_stats}.
	We use the KL divergence between the attention map and a uniform distribution to measure how much the attention distribution diffuses in space. 
	As shown in Figure~\ref{fig: attn_stats}, the higher-resolution attention modules have higher divergence from the uniform distribution, indicating that their attention is less diffused and more concentrated at fine-grained features.
	This verifies the complementarity of low- and high-resolution features as discussed in Section~\ref{sec: multi-scale}.
	
	In addition, we provide a visualization of the learned joint-aware attention in Figure~\ref{fig: vis_attn}(d), which shows intriguing local patterns where the information around the target human joint is emphasized for better 3D human reconstruction.
	Specifically, it has higher weights on the upper and lower sides of the knee, which captures the kinetic status around the joint and helps infer the relative rotation between the thigh and calf.

	Lastly, we employ a self-attention layer $h_\text{self}$ for attention decoupling (Figure~\ref{fig: combine_attn}), which models the target-target dependencies in Figure~\ref{fig: teaser}(c).
	We visualize this layer by showing the interactions between one query joint and all other joints, where the $h_\text{self}$ focuses on neighboring joints on the kinetic tree~\cite{loper2015smpl} to produce more reasonable predictions.

	\vspace{1mm}
	\minisection{Running speed.}
	As shown in Table~\ref{tab:runtime}, the inference of SMPLer and SMPLer-L runs at 96.0 and 73.9 frames per second (fps) respectively, which is effectively real-time.
	While exploiting high-dimensional  multi-resolution features, SMPLer has a nearly three-times running speed and one-fifth GFlops of the baseline Transformers~\cite{lin2021end,lin2021mesh} that are solely based on low-dimensional features, showing the efficiency of the proposed algorithm.
	\vspace{1mm}
	\minisection{Relationship with prior works.}
	The proposed SMPLer is based on two fundamental designs: the attention decoupling and the SMPL-based target representation. 
	These designs substantially distinguish our algorithm from existing Transformers~\cite{lin2021end,lin2021mesh}, which are based on full attention and vertex-based representation and thereby can only use low-resolution features in the attention operation.
	Moreover, the proposed framework also clearly differs from the existing SMPL-based approaches, such as HMR~\cite{kanazawa2018end}, SPIN~\cite{kolotouros2019spin}, and RSC-Net~\cite{xu2021_3d} which are solely based on CNNs and cannot exploit the powerful learning capabilities of Transformers for 3D human shape and pose estimation.
	As a result, they suffer from a large performance gap compared to the state-of-the-arts as evidenced in Table~\ref{tab:compare-h36m-3dpw}.

	The two basic designs naturally lead to the development of several novel modules, including the multi-scale attention and the joint-aware attention.
	While these modules are conceptually simple, to have them work properly in our framework is by no means trivial and requires meticulous algorithmic designs, such as the approach to fuse multi-scale features, the pooling-based positional encoding, the relative positional encoding in the joint-aware attention, and the hierarchical architecture.
	Eventually, with the above designs, SMPLer achieves significant improvement over the state-of-the-art methods~\cite{lin2021end,lin2021mesh} with better efficiency.

	\section{Conclusions}
	We have developed a new Transformer framework for high-quality 3D human shape and pose estimation from a single image.  
	At the core of this work are the decoupled attention and the SMPL-based target representation that allow efficient utilization of high-dimensional features. 
	These two strategies also motivate the design of the multi-scale attention and joint-aware attention modules,
	which can be potentially extended to other areas where multi-scale and multi-scope information is valuable, such as motion estimation~\cite{sun2018pwc} and image restoration~\cite{Nah_2017_CVPR}.
	Meanwhile, the proposed algorithm can also be applied to other 3D reconstruction problems, such as 3D animal reconstruction~\cite{biggs2020wldo} and 3D hand reconstruction~\cite{pavlakos2019expressive} by replacing SMPL with other parametric models, \eg, SMAL~\cite{Zuffi:CVPR:2017} and MANO~\cite{MANO:SIGGRAPHASIA:2017}. 
	Nevertheless, this is beyond the scope of this work and will be an interesting direction for future research.
	
	Similar to existing Transformers, the proposed SMPLer is still a hybrid structure that consists of CNN layers in the backbone. To incorporate attention-based backbones in SMPLer such as \cite{liu2021swin,YuanFHLZCW21} will be another direction worth exploring in future work.
	%
	
	%
	%
	%
	%
	%
	%
	
	%
	%
	%
	%
	%
	%
	
	%
	%
	%
	%
	%
	%
	%
	%
	%
	%
	%
	%
	%
	%
	%
	%
	%
	%

	%
	%
	\ifCLASSOPTIONcaptionsoff
	\newpage
	\fi

	\bibliographystyle{IEEEtran}
	\bibliography{egbib}

\newcommand{\arxiv}[1]{arXiv: #1}
\begin{thebibliography}{100}
\providecommand{\url}[1]{#1}
\csname url@samestyle\endcsname
\providecommand{\newblock}{\relax}
\providecommand{\bibinfo}[2]{#2}
\providecommand{\BIBentrySTDinterwordspacing}{\spaceskip=0pt\relax}
\providecommand{\BIBentryALTinterwordstretchfactor}{4}
\providecommand{\BIBentryALTinterwordspacing}{\spaceskip=\fontdimen2\font plus
\BIBentryALTinterwordstretchfactor\fontdimen3\font minus
  \fontdimen4\font\relax}
\providecommand{\BIBforeignlanguage}[2]{{%
\expandafter\ifx\csname l@#1\endcsname\relax
\typeout{** WARNING: IEEEtran.bst: No hyphenation pattern has been}%
\typeout{** loaded for the language `#1'. Using the pattern for}%
\typeout{** the default language instead.}%
\else
\language=\csname l@#1\endcsname
\fi
#2}}
\providecommand{\BIBdecl}{\relax}
\BIBdecl

\bibitem{lin2021mesh}
K.~Lin, L.~Wang, and Z.~Liu, ``Mesh graphormer,'' in \emph{ICCV}, 2021.

\bibitem{lin2021end}
------, ``End-to-end human pose and mesh reconstruction with transformers,'' in
  \emph{CVPR}, 2021.

\bibitem{bogo2016keep}
F.~Bogo, A.~Kanazawa, C.~Lassner, P.~Gehler, J.~Romero, and M.~J. Black, ``Keep
  it smpl: Automatic estimation of 3d human pose and shape from a single
  image,'' in \emph{ECCV}, 2016.

\bibitem{kanazawa2018end}
A.~Kanazawa, M.~J. Black, D.~W. Jacobs, and J.~Malik, ``End-to-end recovery of
  human shape and pose,'' in \emph{CVPR}, 2018.

\bibitem{kocabas2019vibe}
M.~Kocabas, N.~Athanasiou, and M.~J. Black, ``Vibe: Video inference for human
  body pose and shape estimation,'' in \emph{CVPR}, 2020.

\bibitem{kolotouros2019spin}
N.~Kolotouros, G.~Pavlakos, M.~J. Black, and K.~Daniilidis, ``Learning to
  reconstruct {3D} human pose and shape via model-fitting in the loop,'' in
  \emph{ICCV}, 2019.

\bibitem{kolotouros2019convolutional}
N.~Kolotouros, G.~Pavlakos, and K.~Daniilidis, ``Convolutional mesh regression
  for single-image human shape reconstruction,'' in \emph{CVPR}, 2019.

\bibitem{rajasegaran2021tracking}
J.~Rajasegaran, G.~Pavlakos, A.~Kanazawa, and J.~Malik, ``Tracking people with
  3d representations,'' in \emph{NeurIPS}, 2021.

\bibitem{rajasegaran2022tracking}
------, ``Tracking people by predicting 3d appearance, location and pose,'' in
  \emph{{CVPR}}, 2022.

\bibitem{wang2019re}
J.~Wang, Y.~Zhong, Y.~Li, C.~Zhang, and Y.~Wei, ``Re-identification supervised
  texture generation,'' in \emph{CVPR}, 2019.

\bibitem{xu2021texformer}
X.~Xu and C.~C. Loy, ``3d human texture estimation from a single image with
  transformers,'' in \emph{ICCV}, 2021.

\bibitem{zheng2020pamir}
Z.~Zheng, T.~Yu, Y.~Liu, and Q.~Dai, ``{PaMIR}: Parametric model-conditioned
  implicit representation for image-based human reconstruction,'' \emph{{IEEE
  Transactions on Pattern Analysis and Machine Intelligence}}, 2021.

\bibitem{mir2020learning}
A.~Mir, T.~Alldieck, and G.~Pons-Moll, ``Learning to transfer texture from
  clothing images to {3D} humans,'' in \emph{CVPR}, 2020.

\bibitem{ma2021power}
Q.~Ma, J.~Yang, S.~Tang, and M.~J. Black, ``The power of points for modeling
  humans in clothing,'' in \emph{{ICCV}}, 2021.

\bibitem{alldieck2019learning}
T.~Alldieck, M.~Magnor, B.~L. Bhatnagar, C.~Theobalt, and G.~Pons-Moll,
  ``Learning to reconstruct people in clothing from a single rgb camera,'' in
  \emph{CVPR}, 2019.

\bibitem{alldieck2018video}
T.~Alldieck, M.~Magnor, W.~Xu, C.~Theobalt, and G.~Pons-Moll, ``Video based
  reconstruction of 3d people models,'' in \emph{CVPR}, 2018.

\bibitem{alldieck2019tex2shape}
T.~Alldieck, G.~Pons-Moll, C.~Theobalt, and M.~Magnor, ``Tex2shape: Detailed
  full human body geometry from a single image,'' in \emph{ICCV}, 2019.

\bibitem{alldieck2018detailed}
T.~Alldieck, M.~Magnor, W.~Xu, C.~Theobalt, and G.~Pons-Moll, ``Detailed human
  avatars from monocular video,'' in \emph{{3DV}}, 2018.

\bibitem{li2021ai}
R.~Li, S.~Yang, D.~A. Ross, and A.~Kanazawa, ``Ai choreographer: Music
  conditioned 3d dance generation with aist++,'' in \emph{ICCV}, 2021.

\bibitem{hong2022avatarclip}
F.~Hong, M.~Zhang, L.~Pan, Z.~Cai, L.~Yang, and Z.~Liu, ``Avatarclip: Zero-shot
  text-driven generation and animation of 3d avatars,'' \emph{{ACM Transactions
  on Graphics (SIGGRAPH)}}, vol.~41, no.~4, pp. 1--19, 2022.

\bibitem{sanyal2021learning}
S.~Sanyal, A.~Vorobiov, T.~Bolkart, M.~Loper, B.~Mohler, L.~S. Davis,
  J.~Romero, and M.~J. Black, ``Learning realistic human reposing using cyclic
  self-supervision with 3d shape, pose, and appearance consistency,'' in
  \emph{ICCV}, 2021.

\bibitem{grigorev2021stylepeople}
A.~Grigorev, K.~Iskakov, A.~Ianina, R.~Bashirov, I.~Zakharkin, A.~Vakhitov, and
  V.~Lempitsky, ``Stylepeople: A generative model of fullbody human avatars,''
  in \emph{{CVPR}}, 2021.

\bibitem{peng2021neural}
S.~Peng, Y.~Zhang, Y.~Xu, Q.~Wang, Q.~Shuai, H.~Bao, and X.~Zhou, ``Neural
  body: Implicit neural representations with structured latent codes for novel
  view synthesis of dynamic humans,'' in \emph{CVPR}, 2021.

\bibitem{kwon2021neural}
Y.~Kwon, D.~Kim, D.~Ceylan, and H.~Fuchs, ``Neural human performer: Learning
  generalizable radiance fields for human performance rendering,'' in
  \emph{NeurIPS}, 2021.

\bibitem{chen2022gpnerf}
M.~Chen, J.~Zhang, X.~Xu, L.~Liu, Y.~Cai, J.~Feng, and S.~Yan,
  ``Geometry-guided progressive nerf for generalizable and efficient neural
  human rendering,'' in \emph{{ECCV}}, 2022.

\bibitem{hartley2003multiple}
R.~Hartley and A.~Zisserman, \emph{Multiple view geometry in computer
  vision}.\hskip 1em plus 0.5em minus 0.4em\relax Cambridge university press,
  2003.

\bibitem{vaswani2017attention}
A.~Vaswani, N.~Shazeer, N.~Parmar, J.~Uszkoreit, L.~Jones, A.~N. Gomez,
  L.~Kaiser, and I.~Polosukhin, ``Attention is all you need,'' in
  \emph{{NeurIPS}}, 2017.

\bibitem{devlin2018bert}
J.~Devlin, M.-W. Chang, K.~Lee, and K.~Toutanova, ``Bert: Pre-training of deep
  bidirectional transformers for language understanding,'' in \emph{NAACL},
  2019.

\bibitem{dosovitskiy2020image}
A.~Dosovitskiy, L.~Beyer, A.~Kolesnikov, D.~Weissenborn, X.~Zhai,
  T.~Unterthiner, M.~Dehghani, M.~Minderer, G.~Heigold, S.~Gelly \emph{et~al.},
  ``An image is worth 16x16 words: Transformers for image recognition at
  scale,'' in \emph{ICLR}, 2021.

\bibitem{bosquet2020stdnet}
B.~Bosquet, M.~Mucientes, and V.~M. Brea, ``Stdnet: Exploiting high resolution
  feature maps for small object detection,'' \emph{Engineering Applications of
  Artificial Intelligence}, vol.~91, p. 103615, 2020.

\bibitem{sun2019deep}
K.~Sun, B.~Xiao, D.~Liu, and J.~Wang, ``Deep high-resolution representation
  learning for human pose estimation,'' in \emph{CVPR}, 2019.

\bibitem{fcnn}
J.~Long, E.~Shelhamer, and T.~Darrell, ``Fully convolutional networks for
  semantic segmentation,'' in \emph{CVPR}, 2015.

\bibitem{loper2015smpl}
M.~Loper, N.~Mahmood, J.~Romero, G.~Pons-Moll, and M.~J. Black, ``Smpl: A
  skinned multi-person linear model,'' \emph{{ACM Transactions on Graphics
  (SIGGRAPH Asia)}}, vol.~34, no.~6, p. 248, 2015.

\bibitem{human3.6}
C.~Ionescu, D.~Papava, V.~Olaru, and C.~Sminchisescu, ``Human3. 6m: Large scale
  datasets and predictive methods for 3d human sensing in natural
  environments,'' \emph{{IEEE Transactions on Pattern Analysis and Machine
  Intelligence}}, vol.~36, no.~7, pp. 1325--1339, 2013.

\bibitem{tung2017self}
H.-Y. Tung, H.-W. Tung, E.~Yumer, and K.~Fragkiadaki, ``Self-supervised
  learning of motion capture,'' in \emph{NeurIPS}, 2017.

\bibitem{pavlakos2018learning}
G.~Pavlakos, L.~Zhu, X.~Zhou, and K.~Daniilidis, ``Learning to estimate 3d
  human pose and shape from a single color image,'' in \emph{CVPR}, 2018.

\bibitem{lassner2017unite}
C.~Lassner, J.~Romero, M.~Kiefel, F.~Bogo, M.~J. Black, and P.~V. Gehler,
  ``Unite the people: Closing the loop between 3d and 2d human
  representations,'' in \emph{CVPR}, 2017.

\bibitem{omran2018neural}
M.~Omran, C.~Lassner, G.~Pons-Moll, P.~Gehler, and B.~Schiele, ``Neural body
  fitting: Unifying deep learning and model based human pose and shape
  estimation,'' in \emph{{3DV}}, 2018.

\bibitem{guler2019holopose}
R.~A. Guler and I.~Kokkinos, ``Holopose: Holistic 3d human reconstruction
  in-the-wild,'' in \emph{CVPR}, 2019.

\bibitem{xu2019denserac}
Y.~Xu, S.-C. Zhu, and T.~Tung, ``Denserac: Joint 3d pose and shape estimation
  by dense render-and-compare,'' in \emph{ICCV}, 2019.

\bibitem{jiang2020coherent}
W.~Jiang, N.~Kolotouros, G.~Pavlakos, X.~Zhou, and K.~Daniilidis, ``Coherent
  reconstruction of multiple humans from a single image,'' in \emph{CVPR},
  2020.

\bibitem{zhang2020object}
T.~Zhang, B.~Huang, and Y.~Wang, ``Object-occluded human shape and pose
  estimation from a single color image,'' in \emph{CVPR}, 2020.

\bibitem{zeng20203d}
W.~Zeng, W.~Ouyang, P.~Luo, W.~Liu, and X.~Wang, ``3d human mesh regression
  with dense correspondence,'' in \emph{CVPR}, 2020.

\bibitem{zanfir2020weakly}
A.~Zanfir, E.~G. Bazavan, H.~Xu, W.~T. Freeman, R.~Sukthankar, and
  C.~Sminchisescu, ``Weakly supervised 3d human pose and shape reconstruction
  with normalizing flows,'' in \emph{{ECCV}}, 2020.

\bibitem{song2020human}
J.~Song, X.~Chen, and O.~Hilliges, ``Human body model fitting by learned
  gradient descent,'' in \emph{{ECCV}}, 2020.

\bibitem{Choi_2020_ECCV_Pose2Mesh}
H.~Choi, G.~Moon, and K.~M. Lee, ``Pose2mesh: Graph convolutional network for
  3d human pose and mesh recovery from a 2d human pose,'' in \emph{ECCV}, 2020.

\bibitem{georgakis2020hierarchical}
G.~Georgakis, R.~Li, S.~Karanam, T.~Chen, J.~Ko{\v{s}}eck{\'a}, and Z.~Wu,
  ``Hierarchical kinematic human mesh recovery,'' in \emph{{ECCV}}, 2020.

\bibitem{Moon_2020_ECCV_I2L-MeshNet}
G.~Moon and K.~M. Lee, ``I2l-meshnet: Image-to-lixel prediction network for
  accurate 3d human pose and mesh estimation from a single rgb image,'' in
  \emph{ECCV}, 2020.

\bibitem{zhang2020learning}
H.~Zhang, J.~Cao, G.~Lu, W.~Ouyang, and Z.~Sun, ``Learning 3d human shape and
  pose from dense body parts,'' \emph{{IEEE Transactions on Pattern Analysis
  and Machine Intelligence}}, 2020.

\bibitem{Moon_2022_CVPRW_Hand4Whole}
G.~Moon, H.~Choi, and K.~M. Lee, ``Accurate 3d hand pose estimation for
  whole-body 3d human mesh estimation,'' in \emph{{CVPR Workshops}}, 2022.

\bibitem{li2021hybrik}
J.~Li, C.~Xu, Z.~Chen, S.~Bian, L.~Yang, and C.~Lu, ``Hybrik: A hybrid
  analytical-neural inverse kinematics solution for 3d human pose and shape
  estimation,'' in \emph{{CVPR}}, 2021.

\bibitem{sengupta2021probabilistic}
A.~Sengupta, I.~Budvytis, and R.~Cipolla, ``Probabilistic 3d human shape and
  pose estimation from multiple unconstrained images in the wild,'' in
  \emph{{CVPR}}, 2021.

\bibitem{akhter2015pose}
I.~Akhter and M.~J. Black, ``Pose-conditioned joint angle limits for 3d human
  pose reconstruction,'' in \emph{{CVPR}}, 2015.

\bibitem{zanfir2021neural}
A.~Zanfir, E.~G. Bazavan, M.~Zanfir, W.~T. Freeman, R.~Sukthankar, and
  C.~Sminchisescu, ``Neural descent for visual 3d human pose and shape,'' in
  \emph{{CVPR}}, 2021.

\bibitem{joo2021exemplar}
H.~Joo, N.~Neverova, and A.~Vedaldi, ``Exemplar fine-tuning for 3d human model
  fitting towards in-the-wild 3d human pose estimation,'' in \emph{{3DV}},
  2021.

\bibitem{kolotouros2021probabilistic}
N.~Kolotouros, G.~Pavlakos, D.~Jayaraman, and K.~Daniilidis, ``Probabilistic
  modeling for human mesh recovery,'' in \emph{{ICCV}}, 2021.

\bibitem{dwivedi2021learning}
S.~K. Dwivedi, N.~Athanasiou, M.~Kocabas, and M.~J. Black, ``Learning to
  regress bodies from images using differentiable semantic rendering,'' in
  \emph{{ICCV}}, 2021.

\bibitem{sun2021monocular}
Y.~Sun, Q.~Bao, W.~Liu, Y.~Fu, M.~J. Black, and T.~Mei, ``Monocular, one-stage,
  regression of multiple 3d people,'' in \emph{{ICCV}}, 2021.

\bibitem{zanfir2021thundr}
M.~Zanfir, A.~Zanfir, E.~G. Bazavan, W.~T. Freeman, R.~Sukthankar, and
  C.~Sminchisescu, ``Thundr: Transformer-based 3d human reconstruction with
  markers,'' in \emph{{ICCV}}, 2021.

\bibitem{zhang2021pymaf}
H.~Zhang, Y.~Tian, X.~Zhou, W.~Ouyang, Y.~Liu, L.~Wang, and Z.~Sun, ``Pymaf: 3d
  human pose and shape regression with pyramidal mesh alignment feedback
  loop,'' in \emph{{ICCV}}, 2021.

\bibitem{kocabas2021spec}
M.~Kocabas, C.-H.~P. Huang, J.~Tesch, L.~M{\"u}ller, O.~Hilliges, and M.~J.
  Black, ``Spec: Seeing people in the wild with an estimated camera,'' in
  \emph{{ICCV}}, 2021.

\bibitem{kocabas2021pare}
M.~Kocabas, C.-H.~P. Huang, O.~Hilliges, and M.~J. Black, ``Pare: Part
  attention regressor for 3d human body estimation,'' in \emph{{ICCV}}, 2021.

\bibitem{kanazawa2019learning}
A.~Kanazawa, J.~Y. Zhang, P.~Felsen, and J.~Malik, ``Learning 3d human dynamics
  from video,'' in \emph{CVPR}, 2019.

\bibitem{arnab2019exploiting}
A.~Arnab, C.~Doersch, and A.~Zisserman, ``Exploiting temporal context for 3d
  human pose estimation in the wild,'' in \emph{{CVPR}}, 2019.

\bibitem{sun2019human}
Y.~Sun, Y.~Ye, W.~Liu, W.~Gao, Y.~Fu, and T.~Mei, ``Human mesh recovery from
  monocular images via a skeleton-disentangled representation,'' in
  \emph{{ICCV}}, 2019.

\bibitem{doersch2019sim2real}
C.~Doersch and A.~Zisserman, ``Sim2real transfer learning for 3d human pose
  estimation: motion to the rescue,'' in \emph{NeurIPS}, 2019.

\bibitem{luo20203d}
Z.~Luo, S.~A. Golestaneh, and K.~M. Kitani, ``3d human motion estimation via
  motion compression and refinement,'' in \emph{{ACCV}}, 2020.

\bibitem{choi2021beyond}
H.~Choi, G.~Moon, J.~Y. Chang, and K.~M. Lee, ``Beyond static features for
  temporally consistent 3d human pose and shape from a video,'' in
  \emph{{CVPR}}, 2021.

\bibitem{lee2021uncertainty}
G.-H. Lee and S.-W. Lee, ``Uncertainty-aware human mesh recovery from video by
  learning part-based 3d dynamics,'' in \emph{{ICCV}}, 2021.

\bibitem{wan2021encoder}
Z.~Wan, Z.~Li, M.~Tian, J.~Liu, S.~Yi, and H.~Li, ``Encoder-decoder with
  multi-level attention for 3d human shape and pose estimation,'' in
  \emph{{ICCV}}, 2021.

\bibitem{xu20203d}
X.~Xu, H.~Chen, F.~Moreno-Noguer, L.~A. Jeni, and F.~De~la Torre, ``{3D} human
  shape and pose from a single low-resolution image with self-supervised
  learning,'' in \emph{ECCV}, 2020.

\bibitem{xu2021_3d}
------, ``{3D} human pose, shape and texture from low-resolution images and
  videos,'' \emph{{IEEE Transactions on Pattern Analysis and Machine
  Intelligence}}, 2021.

\bibitem{moreno20173d}
F.~Moreno-Noguer, ``3d human pose estimation from a single image via distance
  matrix regression,'' in \emph{CVPR}, 2017.

\bibitem{rong2021frankmocap}
Y.~Rong, T.~Shiratori, and H.~Joo, ``Frankmocap: A monocular 3d whole-body pose
  estimation system via regression and integration,'' in \emph{{ICCV
  Workshops}}, 2021.

\bibitem{davydov2022adversarial}
A.~Davydov, A.~Remizova, V.~Constantin, S.~Honari, M.~Salzmann, and P.~Fua,
  ``Adversarial parametric pose prior,'' in \emph{{CVPR}}, 2022.

\bibitem{tiwari22posendf}
G.~Tiwari, D.~Antic, J.~E. Lenssen, N.~Sarafianos, T.~Tung, and G.~Pons-Moll,
  ``Pose-ndf: Modeling human pose manifolds with neural distance fields,'' in
  \emph{{ECCV}}, 2022.

\bibitem{cmu_mocap_data}
``{CMU} graphics lab motion capture database,'' \url{http://mocap.cs.cmu.edu/},
  2010.

\bibitem{liu2023sewformer}
L.~Liu, X.~Xu, Z.~Lin, J.~Liang, and S.~Yan, ``Towards garment sewing pattern
  reconstruction from a single image,'' \emph{ACM Transactions on Graphics
  (SIGGRAPH Asia)}, 2023.

\bibitem{liu2021swin}
Z.~Liu, Y.~Lin, Y.~Cao, H.~Hu, Y.~Wei, Z.~Zhang, S.~Lin, and B.~Guo, ``Swin
  transformer: Hierarchical vision transformer using shifted windows,'' in
  \emph{ICCV}, 2021.

\bibitem{carion2020end}
N.~Carion, F.~Massa, G.~Synnaeve, N.~Usunier, A.~Kirillov, and S.~Zagoruyko,
  ``End-to-end object detection with transformers,'' in \emph{ECCV}, 2020.

\bibitem{chen2020pre}
H.~Chen, Y.~Wang, T.~Guo, C.~Xu, Y.~Deng, Z.~Liu, S.~Ma, C.~Xu, C.~Xu, and
  W.~Gao, ``Pre-trained image processing transformer,'' in \emph{{CVPR}}, 2021.

\bibitem{shi2022video}
Z.~Shi, X.~Xu, X.~Liu, J.~Chen, and M.-H. Yang, ``Video frame interpolation
  transformer,'' in \emph{{CVPR}}, 2022.

\bibitem{zhang2019self}
H.~Zhang, I.~Goodfellow, D.~Metaxas, and A.~Odena, ``Self-attention generative
  adversarial networks,'' in \emph{ICML}, 2019.

\bibitem{jiang2021transgan}
Y.~Jiang, S.~Chang, and Z.~Wang, ``Transgan: Two pure transformers can make one
  strong gan, and that can scale up,'' in \emph{{NeurIPS}}, 2021.

\bibitem{ba2016layer}
J.~L. Ba, J.~R. Kiros, and G.~E. Hinton, ``Layer normalization,''
  \emph{\arxiv{1607.06450}}, 2016.

\bibitem{dai2015euler}
J.~S. Dai, ``Euler--rodrigues formula variations, quaternion conjugation and
  intrinsic connections,'' \emph{Mechanism and Machine Theory}, vol.~92, pp.
  144--152, 2015.

\bibitem{jolliffe2016principal}
I.~T. Jolliffe and J.~Cadima, ``Principal component analysis: a review and
  recent developments,'' \emph{Philosophical Transactions of the Royal Society
  A: Mathematical, Physical and Engineering Sciences}, vol. 374, no. 2065, p.
  20150202, 2016.

\bibitem{zhou2019continuity}
Y.~Zhou, C.~Barnes, J.~Lu, J.~Yang, and H.~Li, ``On the continuity of rotation
  representations in neural networks,'' in \emph{{CVPR}}, 2019.

\bibitem{3dpw}
T.~von Marcard, R.~Henschel, M.~J. Black, B.~Rosenhahn, and G.~Pons-Moll,
  ``Recovering accurate 3d human pose in the wild using imus and a moving
  camera,'' in \emph{ECCV}, 2018.

\bibitem{zhang2021temporal}
C.~Zhang, A.~Gupta, and A.~Zisserman, ``Temporal query networks for
  fine-grained video understanding,'' in \emph{{CVPR}}, 2021.

\bibitem{zheng2023potter}
C.~Zheng, X.~Liu, G.-J. Qi, and C.~Chen, ``Potter: Pooling attention
  transformer for efficient human mesh recovery,'' in \emph{{CVPR}}, 2023.

\bibitem{mehta2018single}
D.~Mehta, O.~Sotnychenko, F.~Mueller, W.~Xu, S.~Sridhar, G.~Pons-Moll, and
  C.~Theobalt, ``Single-shot multi-person 3d pose estimation from monocular
  rgb,'' in \emph{3DV}, 2018.

\bibitem{coco}
T.-Y. Lin, M.~Maire, S.~Belongie, J.~Hays, P.~Perona, D.~Ramanan,
  P.~Doll{\'a}r, and C.~L. Zitnick, ``Microsoft coco: Common objects in
  context,'' in \emph{ECCV}, 2014.

\bibitem{mpii}
M.~Andriluka, L.~Pishchulin, P.~Gehler, and B.~Schiele, ``2d human pose
  estimation: New benchmark and state of the art analysis,'' in \emph{CVPR},
  2014.

\bibitem{schonemann1966generalized}
P.~H. Sch{\"o}nemann, ``A generalized solution of the orthogonal procrustes
  problem,'' \emph{Psychometrika}, vol.~31, no.~1, pp. 1--10, 1966.

\bibitem{mixamo}
``Mixamo,'' \url{https://www.mixamo.com/}.

\bibitem{sun2018pwc}
D.~Sun, X.~Yang, M.-Y. Liu, and J.~Kautz, ``Pwc-net: Cnns for optical flow
  using pyramid, warping, and cost volume,'' in \emph{{CVPR}}, 2018.

\bibitem{Nah_2017_CVPR}
S.~Nah, T.~H. Kim, and K.~M. Lee, ``Deep multi-scale convolutional neural
  network for dynamic scene deblurring,'' in \emph{{CVPR}}, 2017.

\bibitem{biggs2020wldo}
B.~Biggs, O.~Boyne, J.~Charles, A.~Fitzgibbon, and R.~Cipolla, ``{W}ho left the
  dogs out?: {3D} animal reconstruction with expectation maximization in the
  loop,'' in \emph{{ECCV}}, 2020.

\bibitem{pavlakos2019expressive}
G.~Pavlakos, V.~Choutas, N.~Ghorbani, T.~Bolkart, A.~A. Osman, D.~Tzionas, and
  M.~J. Black, ``Expressive body capture: 3d hands, face, and body from a
  single image,'' in \emph{{CVPR}}, 2019.

\bibitem{Zuffi:CVPR:2017}
S.~Zuffi, A.~Kanazawa, D.~Jacobs, and M.~J. Black, ``{3D} menagerie: Modeling
  the {3D} shape and pose of animals,'' in \emph{{CVPR}}, 2017.

\bibitem{MANO:SIGGRAPHASIA:2017}
J.~Romero, D.~Tzionas, and M.~J. Black, ``Embodied hands: Modeling and
  capturing hands and bodies together,'' \emph{{ACM Transactions on Graphics
  (SIGGRAPH Asia)}}, vol.~36, no.~6, 2017.

\bibitem{YuanFHLZCW21}
Y.~Yuan, R.~Fu, L.~Huang, W.~Lin, C.~Zhang, X.~Chen, and J.~Wang, ``Hrformer:
  High-resolution transformer for dense prediction,'' in \emph{{NeurIPS}},
  2021.

\end{thebibliography}

\end{document}